\documentclass[pdflatex,sn-mathphys-num]{sn-jnl}

\usepackage{graphicx}%
\usepackage{multirow}%
\usepackage{amsmath,amssymb,amsfonts}%
\usepackage{amsthm}%
\usepackage{mathrsfs}%
\usepackage[dvipsnames]{xcolor}%
\usepackage{textcomp}%
\usepackage{manyfoot}%
\usepackage{booktabs}%
\usepackage{algorithm}%
\usepackage{algorithmicx}%
\usepackage{algpseudocode}%
\usepackage{subfig}
\usepackage{subcaption}
\usepackage{listings}%
\usepackage{ifthen}%
\usepackage{lmodern}%
\usepackage{anyfontsize}%
\usepackage{xspace}
\usepackage{subcaption}

\usepackage{cleveref}%

\newcommand{\task}[1]{{\textsc{{#1}}}}
\def\navigation{\task{Navigation}\xspace}
\def\detection{\task{Detection}\xspace}
\def\tracking{\task{Tracking}\xspace}
\def\ewoo{\task{Eye without optics}\xspace}
\def\ewo{\task{Eye with optics}\xspace}

\newif\ifshowcomments
\showcommentstrue 

\ifshowcomments
    \newcommand{\comment}[1]{\textcolor{olive}{{\em #1}}}
    \newenvironment{multilinecomment}[1]{\begingroup\color{olive}#1}{\endgroup}
    \newcommand{\AY}[1]{\textcolor{blue}{{\em {\bf AY:} #1}}}
    \newcommand{\KT}[1]{\textcolor{RubineRed}{{\em {\bf KT:} #1}}}
    \newcommand{\BC}[1]{\textcolor{red}{{\em {\bf BC:} #1}}}
    \newcommand{\AD}[1]{\textcolor{purple}{{\em {\bf AD:} #1}}}
    \newcommand{\TK}[1]{\textcolor{orange}{{\em {\bf TK:} #1}}}
    \newcommand{\RR}[1]{\textcolor{pink}{{\em {\bf RR:} #1}}}
    \newcommand{\DN}[1]{\textcolor{pink}{{\em {\bf DN:} #1}}}
    \newcommand{\ZT}[1]{\textcolor{brown}{{\em {\bf ZT:} #1}}}
\else
    \newcommand{\comment}[1]{}
    
    \newcommand{\AY}[1]{}
    \newcommand{\KT}[1]{}
    \newcommand{\BC}[1]{}
    \newcommand{\AD}[1]{}
    \newcommand{\TK}[1]{}
    \newcommand{\RR}[1]{}
    \newcommand{\DN}[1]{}
    \newcommand{\ZT}[1]{}
    
\fi

\newif\ifshowoneline
\showonelinetrue 
\showonelinefalse 

\ifshowoneline
    \newcommand{\oneline}[2][show]{%
      \ifthenelse{\equal{#1}{show}}%
        {\noindent {\textcolor{Emerald}{\textbf{One-Liner:} {#2}}}}%
        {\noindent {\textcolor{Emerald}{{#2}}}}%
    }
\else
    \newcommand{\oneline}[2][show]{}
\fi

\newif\ifshowassignments
\showassignmentstrue 

\ifshowassignments
    \newcommand{\assignmentsimple}[1]{{\textcolor{WildStrawberry}{\textbf{Assignment:} {#1}}}}
    \newenvironment{assignment}[1]{\noindent \begingroup\textcolor{WildStrawberry}{\textbf{Assignment:} #1}}{\endgroup}
\else
    \newcommand{\assignmentsimple}[1]{}
    
\fi

\begin{document}

\title[Article Title]{\textit{What if Eye...?} Computationally Recreating Vision Evolution}


\author[1]{\fnm{Kushagra} \sur{Tiwary}\textcolor{Blue}{$^*$}}

\author[1]{\fnm{Aaron} \sur{Young}\textcolor{Blue}{$^*$}}

\author[2]{\fnm{Zaid} \sur{Tasneem}}

\author[1,6]{\fnm{Tzofi} \sur{Klinghoffer}}

\author[1]{\fnm{Akshat} \sur{Dave}}

\author[3]{\fnm{Tomaso} \sur{Poggio}}

\author[4]{\fnm{Dan-Eric} \sur{Nilsson}}

\author[3,5]{\fnm{Brian} \sur{Cheung}\textcolor{BrickRed}{$^{**}$}}

\author[1]{\fnm{Ramesh} \sur{Raskar}\textcolor{BrickRed}{$^{**}$}}

\affil[1]{\orgname{Camera Culture}, \orgdiv{MIT Media Lab}, \orgaddress{\city{Cambridge}, \postcode{02139}, \state{MA}, \country{USA}}}

\affil[2]{\orgname{Computational Imaging Lab}, \orgdiv{Rice University}, \orgaddress{\city{Houston}, \postcode{77005}, \state{TX}, \country{USA}}}

\affil[3]{\orgname{Center for Brains Minds and Machines}, \orgdiv{MIT}, \orgaddress{\city{Cambridge}, \postcode{02139}, \state{MA}, \country{USA}}}

\affil[4]{\orgname{Lund Vision Group}, \orgdiv{Lund University}, \orgaddress{\city{Lund}, \postcode{22100}, \country{Sweden}}}

\affil[5]{\orgname{InfoLab}, \orgdiv{MIT CSAIL}, \orgaddress{\city{Cambridge}, \postcode{02139}, \state{MA}, \country{USA}}}

\affil[6]{\orgname{Charles Stark Draper Laboratory}, \orgaddress{\city{Cambridge}, \postcode{02139}, \state{MA}, \country{USA}}}

\affil[]{\rule{\textwidth}{0.4pt}  \textcolor{Blue}{$^*$} denotes equal authorship, \textcolor{BrickRed}{$^{**}$} denotes equal advising}
\affil[]{Corresponding Authors: \{ ktiwary, cheungb\} @ mit.edu}

\abstract{
Vision systems in nature show remarkable diversity, from simple light-sensitive patches to complex camera eyes with lenses~\cite{land2002AnimalEyes, nilsson_diversity-of-eyes-vision}. While natural selection has produced these eyes through countless mutations over millions of years, they represent just one set of realized evolutionary paths~\cite{fernald2006casting, nilsson2009evolution}. Testing hypotheses about how environmental pressures shaped eye evolution remains challenging since we cannot experimentally isolate individual factors~\cite{nilsson1994pessimistic}. Computational evolution offers a way to systematically explore alternative trajectories~\cite{walter1951machine, holland2003first, Nolfi2016evolutionaryrobotics, winfield2024evolutionary, krause2011interactive}. Here we show how environmental demands drive three fundamental aspects of visual evolution through an artificial evolution framework that co-evolves both physical eye structure and neural processing in embodied agents. First, we demonstrate computational evidence that task specific selection drives bifurcation in eye evolution - orientation tasks like navigation in a maze leads to distributed compound-type eyes while an object discrimination task leads to the emergence of high-acuity camera-type eyes. Second, we reveal how optical innovations like lenses naturally emerge to resolve fundamental tradeoffs between light collection and spatial precision. Third, we uncover systematic scaling laws between visual acuity and neural processing, showing how task complexity drives coordinated evolution of sensory and computational capabilities. Our work introduces a novel paradigm that illuminates evolutionary principles shaping vision by creating targeted single-player games where embodied agents must simultaneously evolve visual systems and learn complex behaviors. Through our unified genetic encoding framework, these embodied agents serve as next-generation hypothesis testing machines while providing a foundation for designing manufacturable bio-inspired vision systems~\cite{lipson2000automatic}. Website: \href{https://eyes.mit.edu/}{eyes.mit.edu}
}

\maketitle

\newpage

\section{Introduction}
\label{sec:introduction}

\begin{figure}[H]
    \centering
    \includegraphics[width=\textwidth]{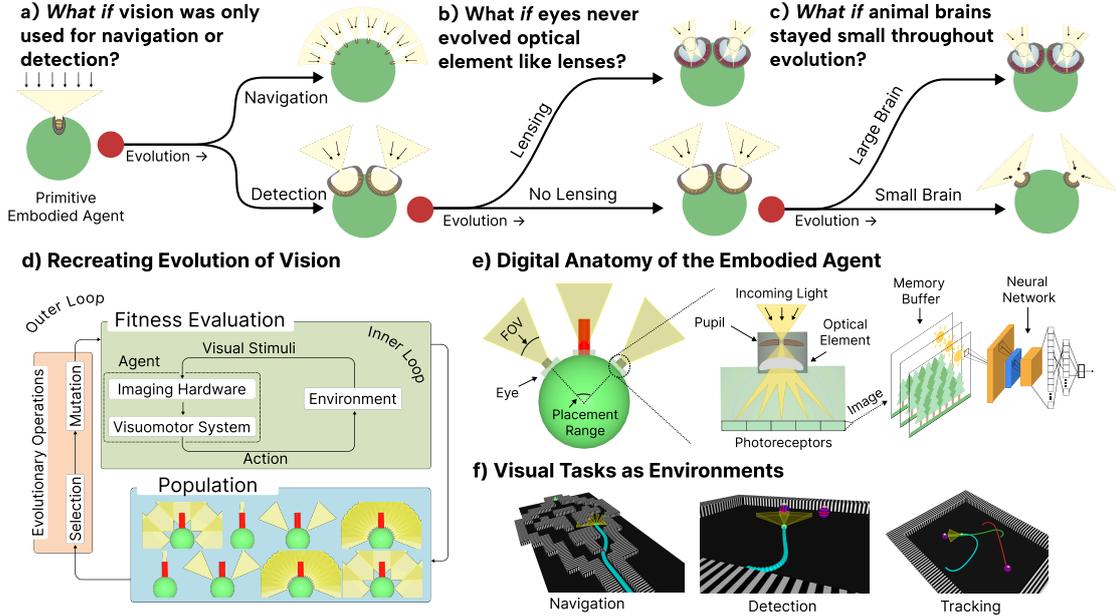}
    \caption{\textbf{Computational evolution of embodied artificial intelligence (AI) agents reveals how environmental pressures shaped natural vision evolution.} We evolve artificial embodied agents to show how three evolutionary branch points shaped vision evolution. We use our framework to understand \textbf{(a)} how environmental specificity led to distinct eye morphologies, \textbf{(b)} how optical elements emerge when embodied agents evolve to discriminate between objects while accounting for physical trade-offs of light throughput vs. spatial precision in an environment, and \textbf{(c)} how visual task error scales as power law with visual acuity and number of parameters revealing that poor visual acuity creates a fundamental bottleneck that cannot be overcome by simply scaling neural capacity. \textbf{(d)} Our framework mirrors natural selection: an outer loop governs genetic inheritance and selection over evolutionary timescales, while an inner loop enables agents to learn through sensory feedback (lifetime adaptation). This nested structure reflects the Baldwin effect~\cite{baldwin1896new}, where lifetime learning can guide and accelerate evolutionary adaptation. \textbf{(e)} The agent's digital anatomy parallels biological visual systems: from eye morphology and placement, through optical elements and photoreceptors (mimicking retinal organization), to neural processing (analogous to visual cortex). \textbf{(f)} Agents are evolved to solve three distinct visual tasks to probe how environmental pressures shape vision: \textit{(i)} \navigation: orientation and obstacle avoidance through a maze-like environment; \textit{(ii)} \detection: object discrimination between a goal object (food) and an adversarial object (poison); \textit{(iii)} \tracking: identical to \detection, but the objects move. Our results highlight how embodied agents can serve as scientific instruments to understand biological visual intelligence.}
    \label{fig:main}
\end{figure}

\oneline[hide]{First paragraph sells the vision: we want to recreate the volution of vision by asking what ifs and enabling the study of counterfactuals; we ask impossible questions since evolution has already happened and its infeasible to study experimentally. We say why you should care about recreating vision, and the technological and scientific impact if we can pull it off. Readers should be convinced that this is an important problem to solve by the end of the paragraph.}

What if vision was only used for navigation or detection? What if eyes never evolved optical elements like lenses? What if animal brains stayed small throughout evolution? Operating over millions of years and culminating in millions of unique perception systems \cite{land2002AnimalEyes}, natural evolution has followed specific evolutionary trajectories in its development of vision. What if there was a tool to instead simulate \textit{alternative} paths that evolution didn't take? By computationally recreating the evolutionary dynamics (i.e., mutation, selection, adaptation) which gave rise to the remarkable diversity of eyes we see today, we can explore different evolutionary trajectories and systematically probe the principles that shape visual diversity. This framework would enable us to test hypotheses about the relationships between eye morphology, neural processing, and environmental pressures, and guide the design of novel vision systems for artificial agents both in nature and engineering. 

\oneline[hide]{Second paragraph introduces the approach: we want to simulate vision evolution in silico. We want readers to be convinced that this `computational emergence of vision' is a new open field.}

In this work, we introduce a framework to elucidate how environmental pressures shaped visual system evolution using embodied artificial intelligence (AI). Our approach evolves the eyes and neural circuitry of embodied agents inside physics-based simulation environments to understand what environmental factors drove vision evolution. While comparative biology has revealed remarkable insight into vision evolution \cite{land2002AnimalEyes, fernald2006casting, cronin2014visual}, testing causal hypotheses or answering ``what-if'' questions has been difficult as these would typically require re-running evolution to understand if the effect still occurs by first performing an intervention on the possible cause \cite{pearl2018book}. Our work builds on two foundational directions. First, pioneered by Grey Walter's machina speculatrix \cite{walter1951machine, holland2003first}, the use of evolutionary robotics to test scientific hypotheses about biological mechanisms and processes \cite{Nolfi2016evolutionaryrobotics, bongard_evo_robotics, krause2011interactive, trianni2014evolutionary, lappalainen2024connectome}. Through learnable embodied agents and a computational model for biological phenomena, we can study predator-prey dynamics \cite{floreano1997adaptive}, brain-body co-evolution \cite{lipson2000automatic}, environmental adaptation \cite{miras2020environmental, ferrante2015evolution, waibel2011quantitative, waibel2009genetic, winfield2024evolutionary}, or study ``why'' questions in neuroscience \cite{kanwisher2023using} such as emergence of foveal image sampling \cite{fovea_cheung} or peripheral vision \cite{pramod2022human}. However, vision evolution has yet to be studied in this context. Second, our work builds on the emergence of deep reinforcement learning (DRL) as a powerful tool for discovery in domains that can be formulated as reward-driven games \cite{silver2017masteringchessshogiselfplay, fawzi2022discovering, silver2018general, mankowitz2023faster}. Our work is the first that illuminates evolutionary principles shaping vision by creating single-player games with specific environmental conditions that embodied agents solve by evolving their vision and learning complex behavior simultaneously. We demonstrate that visually-capable embodied agents trained via DRL can serve as next-generation hypothesis testing machines.

\oneline[hide]{Third paragraph more detailed overview of our approach.}

We first implement a genetic encoding that unifies physical eye morphology, eye optics, and neural processing (\Cref{fig:genotype}), and then computationally mimic the evolutionary process in embodied agents by evolving this genetic encoding to best complete a visual task (\Cref{fig:main}.d). Our framework is the first to computationally recreate vision system evolution -- where complex eyes and behaviors coevolve due to specific environmental pressures. We use this framework to study the emergence of visual capabilities documented across animal phylogeny \cite{fernald2006casting, nilsson2009evolution}. Our encoding integrates morphological, optical, and neural components into a unified genome capable of describing over $10^{20}$ unique configurations, and provides a continuous space for exploring evolutionary pathways (i.e., lens-less cup eyes, camera-type eyes, compound eyes). Subsequently, over generations, agent genes are selected and mutated, leading to the emergence of complex eyes and behaviors for specific visual tasks. This computational survival-of-the-fittest mimics the interplay of variation and selection that shaped biological vision.

\oneline[hide]{Fourth Paragraph about summary of results. Should be sufficiently detailed such that the reader can understand the key insights without reading the rest of the paper.}

Through targeted computational experiments, we establish causal links between specific visual functions \cite{warrant2006invertebrate, nilsson_Visual_Roles, nilsson2009evolution} and solutions, validate aspects of eye evolution as trade-offs between light collection and spatial precision \cite{nilsson1994pessimistic, nilsson2009evolution}, and study relationships between eyes design, neural processing, and visual tasks. Concretely, our scientific contributions are: (1) By strictly changing the visual task an agent is subject to, orientation (\navigation) task vs. object discrimination (\detection) task \cite{nilsson_Visual_Roles, nilsson2009evolution}, we observe a bifurcation in our evolved agents between camera-type and compound-type eyes (\Cref{fig:main}a); (2) We show that the emergence of optical structures, such as focused lensing from primitive simple eyes \cite{nilsson1994pessimistic}, is a key innovation that addresses the fundamental trade-off between light collection and spatial precision (\Cref{fig:main}b); (3) We reveal that sensory acuity and neural capacity scales as a power-law, where decreasing task error requires complimentary improvement in both dimensions --- consistent with observations made in animal vision and AI \cite{kaplan2020scaling,caves2018visual}. Additionally, our engineering contributions are: (1) A framework that evolves vision and behavior of embodied agents through genetic algorithms and deep reinforcement learning in a custom simulation framework; (2) A genetic encoding scheme for vision that describes a diverse set of eyes and cognitive capabilities in addition to being physically-based and realizable. 

\oneline[hide]{Fifth Paragraph about discussion/takeaways. Emphasize the broader impacts.}

Due to biological complexity and computational tractability, we scope our work to recreate the system-level process of vision evolution rather than an imitation of its historical timeline. Since our goal is to understand the mechanistic principles driving vision evolution, we computationally recreate essential elements that shaped natural vision. We model key components universal to biological evolution (\Cref{fig:main}.d), agent's anatomy (\Cref{fig:main}.e), and consult biological studies \cite{srinivasan1996honeybee, nilsson_Visual_Roles} when designing the simulated environments (\Cref{fig:main}.f). Additionally, when studying optical transitions, we use physics-based approximations that are widely-used (\Cref{fig:methods-imaging}), a dynamics engine \cite{mujoco} for interaction, and train our agents only through sensory feedback using reinforcement learning. While our computational framework represents a novel approach to exploring vision evolution, fundamental limitations remain. For example, complex biological phenomena is poorly understood such as eye genomics and development \cite{genetics_eye, gehring1999pax}, bio-physical models \cite{aguirre2019model}, or mechanistic neural circuits \cite{lappalainen2024connectome}. Moreover, modeling evolutionary dynamics is often akin to modeling a chaotic system \cite{doebeli2014chaos} as eye adaptions are often a result of many interdependent pressures. However, our results highlight how embodied agents trained via deep reinforcement learning can serve as scientific instruments to understand biological visual intelligence.

\section{Results}
\label{sec:results}

\oneline[hide]{First, we introduce a simulation framework that allows us to study the evolution of vision in-silico (i.e. the simulation environment). Prior work has focused on individual aspects of evolution, we're doing it all together. We then describe the genetic language and how RL and evolutionary algorithms are combined.}

Our computational framework tests hypotheses about how specific environmental pressures shape eye morphologies, neural architectures, and behavior. While previous work has used evolutionary algorithms to independently design optical systems \cite{10.1117/12.47896,GAGNE20081439}, embodied agent morphologies \cite{gupta2021embodied, sims1991evolving, Nolfi2016evolutionaryrobotics}, or visually-guided behaviors \cite{floreano2004coevolution}, our approach uniquely evolves embodied agents with both eyes and behaviors \textit{together}. This enables the automatic task-driven discovery of diverse vision-based embodied agents. 

\subsection{A \textit{What if ...?} World}
\label{sec:motive-simulation}

\oneline[hide]{First, delve into the simulation world and how we use it to study vision evolution. High level details about the environments we support and implementation details.}

\begin{figure}[t]
    \centering
    \includegraphics[width=\textwidth]{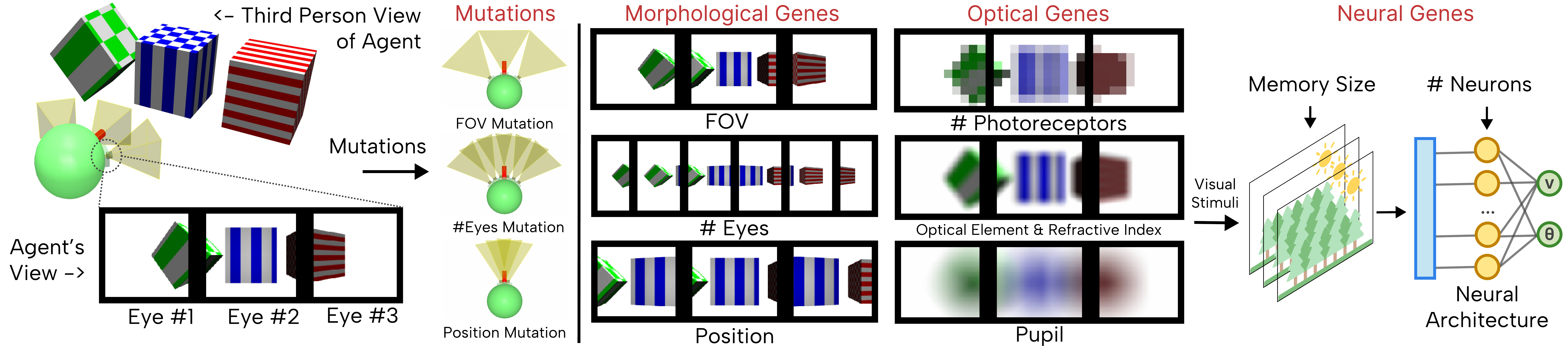}
    \caption{\textbf{Our genetic encoding enables vision to evolve computationally.} Our encoding mirrors the natural separation between sensory and neural development through three gene clusters. \emph{Morphological genes} determine agent properties relating to spatially sampling the environment such as eye placement and field of view. \emph{Optical genes} determine agent properties relating to how each eye interacts with incoming light in a physically plausible way such as \# photoreceptors, optical elements, pupil size). \emph{Neural genes} describe the behavior learning capacity of the agent. These independently mutable genes enable the computational exploration of evolutionary pathways that mirror those in natural vision evolution.}
    \label{fig:genotype}
\end{figure}

We model the world in which embodied agents interact as a deep reinforcement learning environment, where agents must evolve appropriate vision capabilities such that they can learn effective behavior. It's infeasible, however, to model \textit{all} factors which contribute to the evolution of vision; thus, we model each environment as corresponding to a specific \textit{function} of vision, such as orientation or object discrimination. In this way, each environment represents a single task which models the functional pressures hypothesized to garner the emergence of vision. We focus on modeling three distinct tasks to isolate their effects on vision evolution ~\cite{nilsson_Visual_Roles,nilsson_diversity-of-eyes-vision}: \navigation, \detection, and \tracking. We create these tasks in a MuJoCo simulation environment ~\cite{towers2024gymnasium, mujoco} with custom changes to support complex imaging models and evolutionary search. Each agent in this environment is modeled as a point mass (the green sphere in \Cref{fig:genotype}) with a heading and forward velocity. For more technical details, please see \Cref{sec:methods}. 

\subsection{Genetic Encoding for Vision}
\label{sec:motive-design-space}

\oneline[hide]{Overview as to why we need a genetic language and then how we use it.}

Vision in nature has co-evolved as a function of sensing and the underlying neural circuitry; subsequently constraining behavior that an animal learns during its lifetime \cite{nilsson_diversity-of-eyes-vision,baden2024evolution, nilsson2009evolution}. At a population scale, this continuous feedback loop between evolution and learning ensures the Baldwin Effect \cite{baldwin1896new} where learned behaviors affect the evolution and selection of genetic traits of future generations. Similarly, we create a genotype that directly encodes both the physical (morphological and optical) and the neurological components of an agent's vision. Rather than incorporating neural network weights directly in the genomic encoding \cite{neat, salimans2017evolution, conti2018improvingexplorationevolutionstrategies}, we train each agent from scratch to learn behaviors specific to the mutated eye design.

\oneline[hide]{More technical details on the exact genetic language we use.}

Similar to nature, our genetic encoding scheme needs to be general enough to allow the emergence of a diverse set of eyes and cognitive capabilities while being physically realizable. Therefore, we first conceptualize an agent's eye as a physical sensor that converts photons into neural impulses. These impulses are then followed by cognition, which enables agent's to interact within the environment. We categorize the genotype into three sub-genes (\Cref{fig:genotype}): morphological genes that control morphological features such as placement, quantity and field of view, optical genes that controls features about the captured image, and neural genes that controls the agent's behavior. Each gene describes a subset of an agent's visual system and is mutated through specific evolutionary operators (\Cref{fig:genotype}). The exact genetic encoding is discussed in more detail in \Cref{sec:methods-genotype}. 

\subsection{Co-Evolution of Vision and Behavior}

\oneline[hide]{We then introduce the idea of the dual loops we use, along with high level details about how we do evo search and RL. We save specifics of genetic language for next subsection.}

Our approach computationally simulates nature's co-evolution approach to vision innovation: changes in sensory capabilities directly influence behavioral performance, which in turn guides the evolution of future eye morphologies, optics and behaviors. We implement co-evolution through two nested loops that mirror the interplay between evolutionary timescales and lifelong adaptation. Over generations, the outer evolutionary loop utilizes the Covariance Matrix Adaptation Evolutionary Strategy (CMA-ES) \cite{hansen1996adapting, nevergrad} to enable efficient selection and mutation of populations of agents. Within each generation, the inner learning loop, an agent with the selected genotype is instantiated and trained to solve a visual task through reinforcement learning via Proximal Policy Optimization (PPO) \cite{schulman2017proximal}. During the agent's lifetime, the agent's performance is evaluated within the same visual task. The fitness of each agent is then used in the following generation to selectively adapt the populations genotypes. We discuss the evolution and learning loops in more detail in \Cref{sec:methods}.

\subsection{What if the goals for vision were different?}
\label{sec:results-env}

\begin{figure}[ht!]
    \centering
    \includegraphics[width=\textwidth]{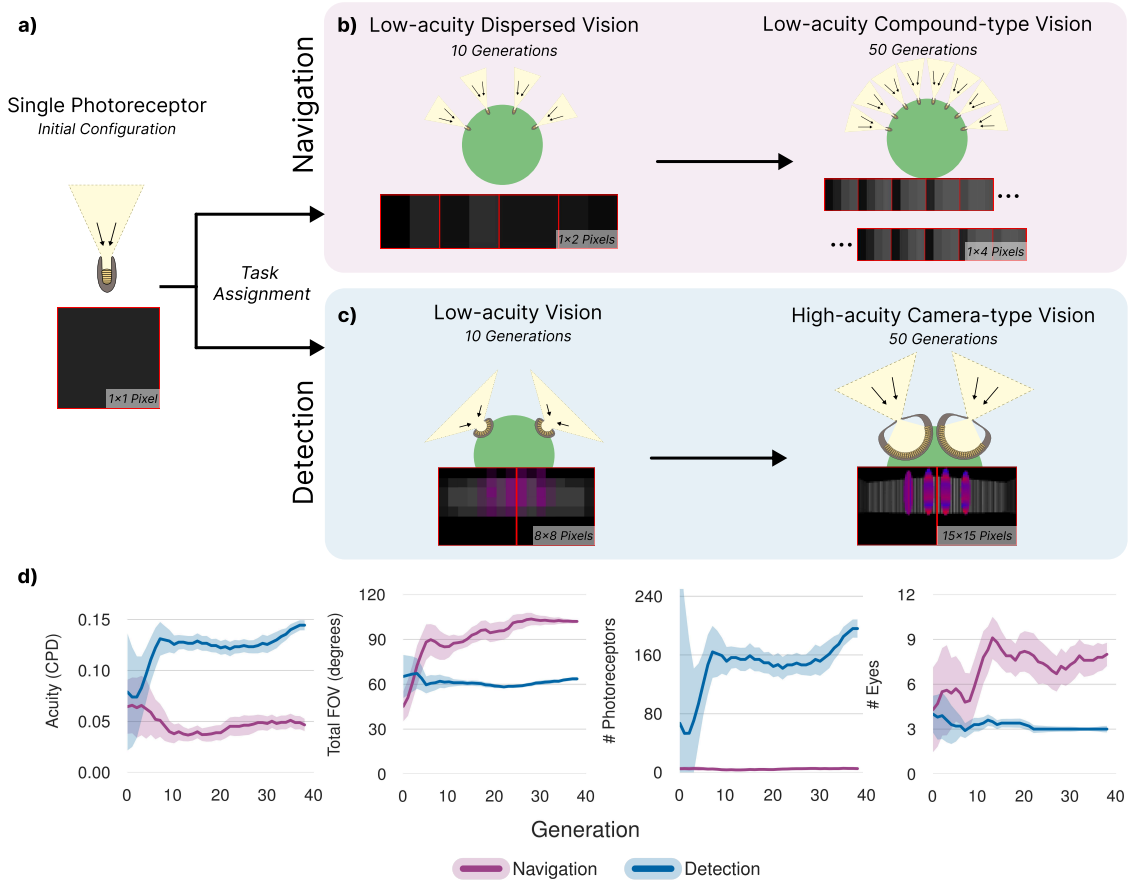}
    \caption{\textbf{Low- and high-acuity spatial tasks lead to compound and camera eyes, respectively.} \textbf{(a)} We initialize a population of agents for two visual tasks (\detection and \navigation) with a single eye with one photoreceptor. We then evolve a population of agents subject to morphological mutations: add photoreceptor, add eye, and adjust placement. In the \navigation task, we first observe an emergence of dispersed vision, where many eyes are employed. By 50 generations, a compound-type eye emerges; that is, a vision system consisting of 10 individual eyes, each with 16 photoreceptors (4 $\times$ 1 resolution), distributed over the entire diameter of the agent. \textbf{(c)} In the \detection task, we initially observe the emergence low-resolution vision. After 50 generations, the population has converged on a morphology consisting of two forward facing, high-resolution camera-type eyes each with 225 photoreceptors (15$\times$15 resolution). \textbf{(d)} Configuration vs generation plots are shown, depicting the evolutionary progression of the mean agent in the population and the task dependence on evolutionary adaptation. The plots show the mean and 95\% bootstrapped confidence interval, respectively.}
    \label{fig:camera-compound}
\end{figure}

\oneline[hide]{Intro to/motivate environmental adaptation, specifically camera-type and compound-type eyes.}

Understanding how specific visual tasks shaped the evolution of eyes remains a major challenge because animals are required to solve multiple visual tasks simultaneously. For instance, honeybees have evolved compound eyes with around 5,000 individual receptors, balancing trade-offs between extracting optic flow to maintain equidistance from obstacles and regulate flight speed, and sufficient spatial resolution to discern body movements of other bees in their colony \cite{srinivasan1996honeybee}. This coupling of tasks in nature makes it difficult to understand how individual visual demands influence eye evolution. For instance, dragonflies have evolved  compound eyes with high-resolution regions, making it challenging to identify which visual adaptation was a result of which environmental pressure. Thus, would evolution converge on similar eye morphologies as found in nature if we could isolate individual visual tasks? To address this, we create two distinct environments that isolate specific visual demands, allowing us to observe how eye morphologies evolve when optimizing for a single task.

\oneline[hide]{Introduce experiment design. Highlight key results and takeaways}

Our computational framework tests vision evolution through two distinct tasks that isolate different environmental pressures. The \navigation task is a goal-less orientation task (\cite{nilsson_Visual_Roles}) where agents are incentivized to traverse a maze environment as fast as possible while avoiding collisions with walls and forward barriers which are alternating with white/black striped patterns of different frequencies (similar to navigational setups that test navigational abilities within honeybees \cite{srinivasan1996honeybee}). Conversely, the \detection task is an object discrimination task where agents choose the goal sphere between three visually similar spherical objects in an open environment (this can be conceptualized as identifying food from poison with the only difference being the rotation of a high frequency spherical pattern on the sphere). In both tasks, agents control only their forward speed and heading. Agents in both tasks are initialized by generating a population via randomly mutating the genotype of a primitive agent (a single eye with one photoreceptor and a field-of-view of $45^{\circ}$). Over the course of evolution, agents mutate by using only morphological constraints: adding or remove photoreceptors, adding or removing eyes, and adjusting the eye's placement. As the \navigation task is essentially two-dimensional, only photoreceptors along the horizontal width of the eye are added/removed. We compare the resulting imaging systems of the optimized agents using cycles per degree (CPD)~\cite{caves2018visual, burton2008scaling, brooke1999scaling}, which is a measure of the spatial frequencies observable to the agent; a higher cpd value corresponds to a better ability resolve fine spatial details in a scene~\cite{caves2018visual} (we describe CPD in more detail in \Cref{app:design-space}).

\oneline[hide]{Eye Result and Takeaway}

From an initial configuration of one eye with a single photoreceptor, agents evolve distinctly different morphologies under each task. To enforce realistic physical constraints, we limit the allowed configurations to prevent overlap based on photoreceptor size and eye radius. Similar to flying insects that navigate complex environments at high speeds \cite{srinivasan1996honeybee}, our \navigation-specialized agents evolved compound-type eyes with 10 widely distributed visual units, each containing 4 photoreceptors (1$\times$4) (\Cref{fig:camera-compound}b). This configuration optimizes full-body coverage with a $135^{\circ}$ total field-of-view, enabling rapid environmental sampling during high-speed maze traversal. In contrast, \detection-specialized agents develop two forward-facing eyes with 225 photoreceptors (15$\times$15) each (\Cref{fig:camera-compound}c), concentrating their visual resources frontally. This evolutionary divergence reflects task-specific optimization: navigation agents maximize spatial awareness through distributed low-resolution sensing, while detection agents sacrifice peripheral vision for enhanced frontal acuity, enabling object discrimination at greater distances within their fixed time constraint.

\oneline[hide]{Neural Network Result and Takeaway}

Our agent's morphological divergence directly influences the topology of their neural network used to learn task behavior. A compound eye configuration in our genetic scheme enables parallel processing of visual information --- each additional eye gets its own visual processing unit (MLPs) that handles processing of a specific part of the visual field before the low dimensional features from each eye are concatenated. Conversely, in the \detection task, agents evolve camera-type vision (2-3 forward facing eyes) with with larger input arrays ($15\times15\times3$) per eye.

\oneline[hide]{Highlight the interesting stuff and relate it to nature.}

Our computational isolation of visual tasks reveals fundamental patterns that parallel natural evolution \cite{land2002AnimalEyes,hansen1996adapting,wehner2003desert,jeffery2009regressive,hogg2014arctic}. The emergence of distinct morphologies from identical starting conditions demonstrates how environmental demands can drive visual specialization. Like bees using wide-field motion detection \cite{srinivasan1996honeybee}, our navigation agents evolved distributed sensing for efficient environmental sampling. When quantifying visual capabilities using CPD measurements \cite{caves2018visual, burton2008scaling, brooke1999scaling}, we found a clear trade-off between spatial coverage and resolution that mirrors natural systems (\Cref{fig:camera-compound}d). This trade-off manifests across species \cite{cronin2014visual}, where animals evolve either enhanced acuity or broader fields of view based on their ecological needs. Our simulation results provide counterfactual evidence that task-specific selection pressures can drive the emergence of these distinct visual and neural processing strategies. 

\subsection{What if eyes could bend light?}
\label{sec:results-optics}

\begin{figure}[t]
    \centering
    \includegraphics[width=\textwidth]{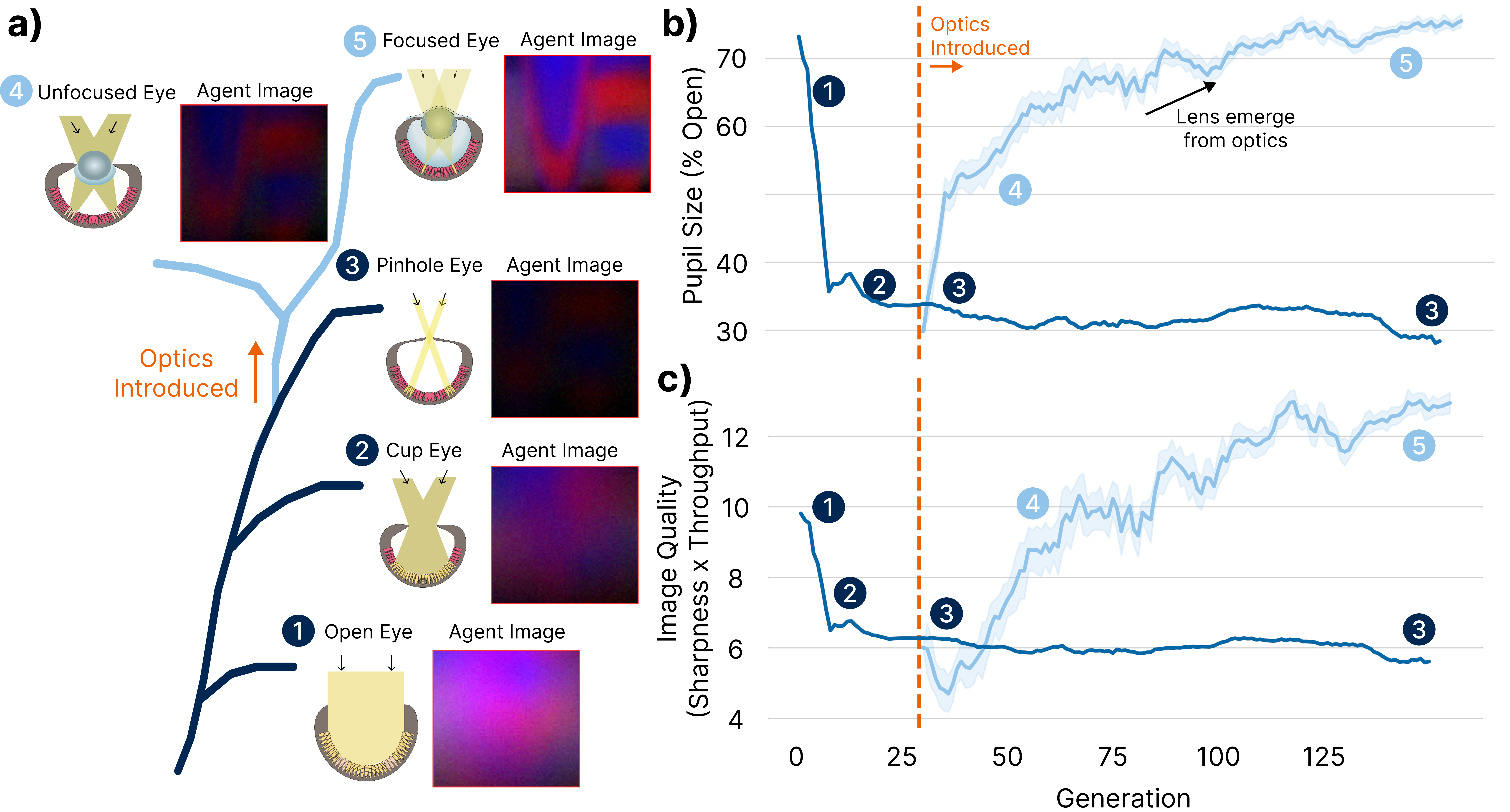}
    \caption{
    \textbf{Computational evolution reveals how lensing resolves a fundamental trade-off in vision.} We demonstrate that to achieve maximum fitness in the \detection task, evolution learns to evolve optical structures against two competing objectives: achieving high spatial precision and maximizing light collection. \textbf{(a)} The evolutionary sequence shows five key stages of eye development: (1) open pupil with maximum light collection but poor spatial precision, (2) cup eye and (3) pinhole eye that achieve better spatial precision by reducing pupil size at the cost of light collection, followed by the emergence of (4) unfocused and (5) focused lens-based eyes that maintain spatial precision by evolving optical structures while allowing larger pupils for more light collection. Agent images show the scene as perceived at each stage. \textbf{(b)} Without optics (dark blue), pupil size decreases to improve precision, sacrificing the signal-to-noise ratio (SNR). When lensing is enabled (orange line, generation 30), larger pupils emerge as lenses are evolved maintain precision while increasing light throughput. \textbf{(c)} The Image Quality metric (image sharpness × light throughput) quantifies this trade-off resolution: pinhole eyes (3) plateau at low values due to limited light collection, while lens-based eyes (4,5) achieve higher quality by combining good spatial precision with larger pupils. This mirrors the evolutionary pressure that drove the emergence of biological lenses, which enabled enhanced vision across lighting conditions.
    }
    \label{fig:pupil}
    
\end{figure}

\oneline[hide]{Introduction to lensing and experimental design.}

Early visual systems faced a fundamental trade-off between light collection and acuity, progressing from simple light-sensitive patches to cup-shaped eyes with smaller apertures \cite{schwab2018evolution, nilsson2013eye}. While decreasing the size of the aperture and creating pinhole-like designs is a straightforward way to improve image formation, they severely limit light collection. This trade-off creates a performance ceiling, where further improvements in spatial resolution through pinhole designs are limited by the lack of light. This inherent limitation ultimately restricts the visual capabilities of such systems, causing a saturation in performance. We see this manifested in our results where agents with pinhole eyes plateau in fitness and are not able to achieve the performance benefits of improved spatial resolution. What if we introduced optical elements that can redirect light into our agents? Lenses emerged as a innovation in biological evolution and we investigate the impact of enabling optical elements capable of bending light in our framework.

To isolate optical evolution, we restrict mutations to the optical subspace: pupil size, optical element, and refractive index. We fixed the remaining morphological genes to the parameters evolved in the \detection task in \Cref{fig:camera-compound}. Pupil size controls the signal-to-noise (SNR) ratio by controlling the total light throughput on the retina; since the noise in the environment is fixed, SNR decreases as pupil size decreases. The optical element is represented as a 2D array that can be programmed into lenses of different shapes (modeled as a diffractive optical element (DOE)) \cite{martel2020neuralsensors, wu2019phasecam3d, e2e_sitzmann}. Refractive index controls the bending of light within the optical element. These three parameters are general enough to be represent a large number of different shapes. We discuss these parameters, the physics-based rendering model, and their relation to real eyes in \Cref{sec:methods}. 

We set out to explore how introducing optical elements might change the evolutionary trajectory of our agents’ vision in the \detection task. To do this, we designed two distinct experiments. In the first, \ewoo, we allowed only the pupil size to mutate while keeping other factors fixed. This setup mirrored the early stages of eye evolution, where organisms could merely adjust their aperture to balance light collection and spatial resolution. Across generations, a clear progression emerged: agents began with fully open apertures (\Cref{fig:pupil}.1) that captured plenty of light but produced blurry images; they next converged on cup-shaped eyes (\Cref{fig:pupil}.2), trading some brightness for reduced blur; and ultimately, they settled on near-pinhole eyes (\Cref{fig:pupil}.3) that offered sharper vision at the cost of even lower light throughput.

After 30 generations in this baseline setting, we initiated a “counterfactual” experiment, \ewo, where agents could also mutate their optical element and refractive index. The sudden introduction of random lens shapes initially caused a dip in performance. However, between generations 30 and 50, evolution steadily refined these diffuse structures into slightly convex shapes (\Cref{fig:pupil}.4), showing the first signs of lens-like focusing. Beyond generation 50, agents reached a turning point: the optical element evolved into well-defined lenses with symmetric point spread functions, while still retaining relatively large pupils (\Cref{fig:pupil}.5). As a result, these agents escaped the tight trade-off imposed by the pinhole design—maintaining a crisp image while collecting more light—leading to higher fitness and more consistent goal discrimination. We quantify these improvements with an “Image Quality” metric, defined as the product of spatial precision (from the Modulation Transfer Function of the point spread function) and maximum light throughput. For further details on the fitness trajectories, PSNR/SSIM analyses, and 3D visualizations of the final lens designs, we refer to Appendix \Cref{app:lensing_analysis}.

\oneline[hide]{Detail results from \ewoo: no lensing enabled.}


\oneline[hide]{Detail results from \ewo: lensing enabled.}


Additionally, \ewoo and \ewo fitness trajectories (\Cref{fig:supp_fitness}) reveal a stark contrast in detection capabilities: while \ewoo agents achieve only sporadic goal detection through random encounters, \ewo agents evolve reliable detection strategies by generation 130, with top performers consistently identifying multiple goals among the adversarial objects. \Cref{fig:supp_fitness} also plots immediate and steady improvements in both PSNR and SSIM. \Cref{fig:supp_optical_elements} shows how high-performing agents (F$>$24.4) evolved singular peaks and smoother optical responses in their optical elements, while poor performers (F$<$3.4) developed fragmented, and multi-peaked patterns. This is also reflected in the dispersed and compact PSFs of low and high performing agents respectively \Cref{fig:supp_mtf}.

\oneline[hide]{Provide context and key takeaways as to why this section is interesting.}

Our results demonstrate a critical sequence that illuminates why lenses emerged over the course of vision evolution. While our agents were simply tasked with discriminating between similar-looking objects, it required the populations to evolve effective eyes subject to the fundamental trade-off in vision evolution: balancing spatial precision (needed to discriminate similar objects) with light collection (needed for reliable vision). Our Image Quality metric, which combines MTF-derived spatial precision with light throughput, quantifies this trade-off directly. In \ewoo, where only pupil size could vary, spatial precision saturates as pinhole eyes sacrifice light collection for acuity, resulting in dark, noisy vision. \ewo reveals how lens evolution resolves this constraint: DOEs evolve into lenses that maintain spatial precision while allowing larger pupils, eliminating the precision vs. light throughput trade-off. Our counterfactual experiment suggests that without this innovation, accurate vision would have been restricted to high-light conditions; for example, while pinhole eyes can technically produce sharp images, they are rarely found in nature due to the loss in light throughput. This aligns with proposed explanations which suggest that lens development coincided with an increase in eye size \cite{schwab2018evolution, nilsson2013eye}. The lens thus represents a fundamental innovation in the evolutionary solution space, discovered by our agents not through direct optimization of optical properties, but through the demands of achieving accurate perception on a specific behavioral task.

\subsection{What if the brain became larger?}
\label{sec:results-neural}

\begin{figure}[ht!]
    \centering
    \includegraphics[width=\textwidth]{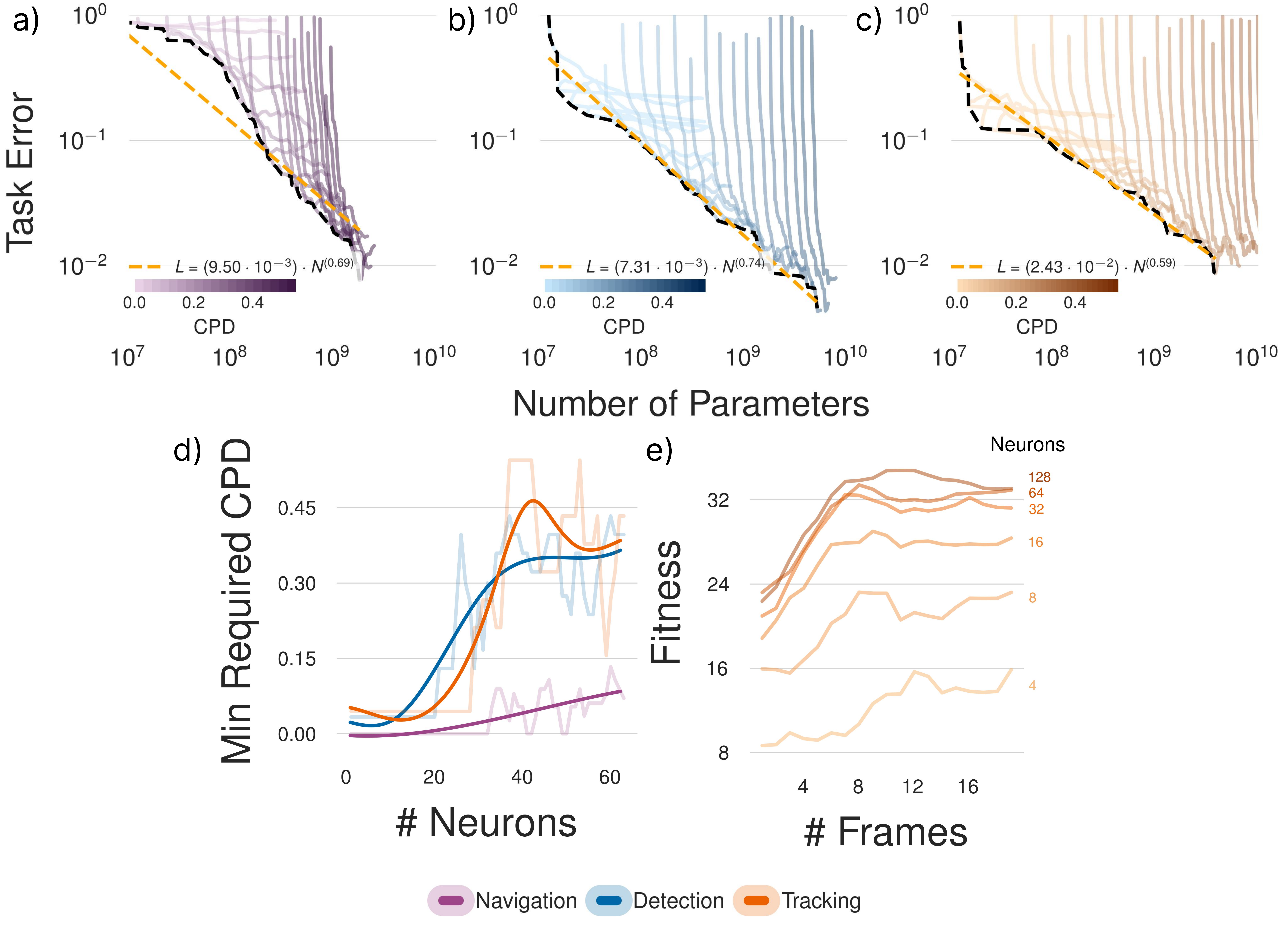}
    \caption{
    \textbf{Task-dependent scaling laws reveal how sensory acuity bounds performance and how temporal memory compensates for neural capacity} \textbf{(a-c)} Our experiments reveal visual task-dependent power law scaling between number of parameters and sensory acuity (CPD). This demonstrates that scaling in sensory input is required for embodied tasks to avoid a bottleneck that cannot be overcome by neural scaling alone. \textbf{(d)} A minimum required visual acuity compared to number of parameters for different embodied tasks suggest a hierarchy in the emergent behaviors depending on the task. \textbf{(e)} Temporal processing shows complementary scaling with neural capacity, where increased temporal memory (\# Frames) can compensate for reduced neural processing, which is particularly evident in tasks with larger networks. Together, these results quantify how visual intelligence emerges from the interplay between sensory, neural, and temporal capabilities.
    }
    \label{fig:neural}
\end{figure}

Biological visual intelligence emerges from the interplay and scaling between sensory hardware, morphology, and neural processing \cite{2018AllometryTS, venditti2024co}. While artificial intelligence relies on fixed sensors (RGB cameras) and scales with the number of parameters, nature has evolved diverse eye-brain systems that scale in complexity together to solve intelligent tasks. By varying eye acuity (cycles-per-degree) \cite{CAVES2018358}, neural network size, and temporal processing \cite{abrams2014trade}, we investigate how these resources shape the evolution of visual intelligence and task-specific performance in embodied agents.

Our analysis reveals distinct power law scaling between task performance and neural capacity across navigation, detection, and tracking tasks. Performance across network sizes (\Cref{fig:neural}.a-c) follows characteristic power laws ($L = (9.50 \cdot 10^{-3}) \cdot N^{0.69}$ for navigation, $L = (7.31 \cdot 10^{-3}) \cdot N^{0.74}$ for detection, and $L = (2.43 \cdot 10^{-2}) \cdot N^{0.59}$ for tracking). This power law defines a predictable improvement in task error as a function of increasing network size. But this trend is only persists when another quantity, the level of visual acuity, is also able to improve. Each acuity level bounds the maximum achievable task performance (minimum task error). Low acuity models hit performance ceilings, demonstrating that poor visual acuity creates a fundamental bottleneck that cannot be overcome by simply scaling neural capacity. This resource limitation mirrors the power laws for scaling (also known as `scaling laws') seen in artificial intelligence systems, where performance is bounded by the interplay of model size, computational resources, and data availability \cite{kaplan2020scaling}.

These scaling relationships reveal how evolving morphological constraints like eye structure and number of neurons (brain size) act as fundamental resources that can affect scaling in embodied agents performance. Sensory acuity (measured with CPD) is also a resource which limits the throughput of information from the scene to agent based on fundamental limitations of light transport. Our results demonstrate that power law scaling only holds when both acuity and parameter resources scale appropriately, with performance saturating when either becomes a bottleneck (i.e. in the relationship between visual acuity and number of parameters increasing model size cannot overcome fundamental sensing limitations). However, biological evolution has repeatedly overcome such constraints through scaling across different genetic traits \cite{2018AllometryTS, venditti2024co} - suggesting parallel opportunities for artificial systems where scaling of data and parameters alone may be insufficient without corresponding scaling in sensory capabilities.

Critically, we also identify transition points that reveal fundamental limits in visual processing (\Cref{fig:neural}.d). For a fixed neural network size, increasing sensory acuity (CPD) beyond certain thresholds yields diminishing returns, with each task showing distinct saturation points. For example, the detection task requires higher minimum acuity to achieve comparable performance, suggesting a hierarchy in the task emergence and visual processing demands of different behaviors.

For time-oriented tasks like tracking, we uncover a compensatory relationship between neural processing and temporal memory (\Cref{fig:neural}.e), where increased temporal information can offset reduced neural processing capacity and vice versa. This demonstrates that equivalent task performance can be achieved either through sophisticated processing of individual frames or through extended temporal integration of simpler visual features.

\section{Discussion}
\label{sec:discussion}

\oneline[hide]{First paragraph reminds what the claims are and how our results verify those claims.} 

Similar to natural evolution, we follow a \textit{function over form} approach, where we code the desired function through fitness and let evolutionary search discover a variety of forms that are optimal for the fitness. This results in our agent's form (eye design and learned behavior) to emerge solely from functional pressures from the environment such as orientation, obstacle avoidance, or object discrimination. The emergent features resemble principles of real biological evolution. These results affirm our central claim that embodied agents trained with reinforcement learning can serve as hypothesis-testing machines for vision and vision evolution. The evolutionary outcomes we present are a result of the co-evolution of vision- hardware (physical eye morphology and structure) and software (learned behavior of the agent). Lastly, in our current approach, we evolve our agents under isolated environmental pressures i.e. cases where agents are heavily biased to evolve to solve a single task. However, in the natural world animals have evolved to jointly solve diverse tasks found in their ecological niches. While our framework can be easily extended for diverse visual tasks, isolated scenarios help us understand the extreme cases. 

Since our work is the first in this space, our results point towards open technical challenges that will enable a wider variety of hypothesis to be tested. For instance, future research can be extended to incorporate explore multi-agent interactions where multiple species evolve in shared environments, applying gradient-based methods, or incorporating richer light properties like spectral, polarization, or temporal sampling. Additionally, future work could include incorporating bio-physical models of vision \cite{aguirre2019model} or replace neural networks with mechanistic circuits derived from fly connectomes \cite{lappalainen2024connectome, shinomiya2022neuronal}.

\oneline[hide]{Second paragraph discusses technical merits and what are the next few papers that people will write in this space.}

Our framework provides a discovery tool by enabling large scale computational evolution of vision in embodied artificial agents. For biologists and cognitive scientists, this approach allows systematic manipulation of key variables to test alternative hypothesis or counterfactuals --- such as isolating the effects of optical elements from neural processing, or testing how specific environmental pressures drive eye morphology. Much like natural and artificial evolution \cite{lehman2019surprisingcreativitydigitalevolution}, our framework also demonstrates remarkable creativity in discovering solutions --- for instance, it independently evolved compound-eye architectures without being explicitly designed to do so. For engineers, these evolutionary simulations reveal design principles for artificial vision systems, particularly valuable when optimizing for practical constraints like energy efficiency and manufacturability \cite{diser2023, Tiwary2024Roadmap, lipson2000automatic}.

\section{Methods}
\label{sec:methods}

\noindent \textbf{Learning loop.} 
The learning loop is the mechanism for which we score each agent. Via reinforcement learning, we train the brain of the agent (i.e., neural network parameters). Reinforcement learning serves as a mechanism for learning representations of the environment through interactions with it. The subsequent score, or fitness, of the agent is determined from the average reward it receives over six evaluation episodes after training. We utilize an open-source implementation of the Proximal Policy Optimization (PPO) algorithm \cite{stable-baselines3,schulman2017proximal}. Each agent is trained up to 1 million total steps, though training may be terminated early if no improvement is found after five evaluations.

Reinforcement learning algorithms have been shown to have a strong dependence on the random seed used to initialize the environment \cite{henderson2018deep}. Thus, during the evolution loop, we allow the same agent genotype to be sampled multiple times. Additionally, each agent's training loop is initialized with a unique random seed such that configurations sampled with the same genotype are not subject to the same seed. This allows for a more robust evaluation of the agent's performance. 

\bigskip \noindent \textbf{Observations.} 
An agent interacts with its environment through actions based on its observations. The observations are created by compositing the images captured from each eye. For example, if an agent's vision system consists of 5 eyes with four photoreceptors in each eye, the resulting observation by the full agent ``eye'' at each time step will be a tensor of size 5 x 4 x 1 x 3. Furthermore, for each eye, the previous observations are stacked in a memory buffer. If the memory buffer is of size 10, then a tensor if size 10 x 5 x 4 x 1 x 3 is provided to the underlying agent network. In addition to the visual stimuli, we provide a single boolean that describes whether an agent is in physical contact with an object at the current time step and the previous action that was taken. Although not needed for the agent to solve the tasks, we have found that providing contact information and previous action as observations led to convergence nearly twice as fast; in an evolutionary search context, this speed-up significantly improves the overall optimization time.

\bigskip \noindent \textbf{Reward function.} The reward function is used in RL to drive policy optimization towards some desired observation and action mapping. In our case, each task has a unique reward function:

\begin{equation}
    \quad R_{\textsc{Navigation}} = \phantom{-}\lambda \left( \|\mathbf{x}_t - \mathbf{x}_0\| - \|\mathbf{x}_{t-1} - \mathbf{x}_0\| \right) + w_g + w_c 
\end{equation}
\begin{equation}
    \quad R_{\textsc{Detection}} = -\lambda \left( \|\mathbf{x}_t - \mathbf{x}_f\| - \|\mathbf{x}_{t-1} - \mathbf{x}_f\| \right) + w_g + w_a + w_c 
\end{equation}
\begin{equation}
    \quad R_{\textsc{Tracking}} = -\lambda \left( \|\mathbf{x}_t - \mathbf{x}_f\| - \|\mathbf{x}_{t-1} - \mathbf{x}_f\| \right) + w_g + w_a + w_c
\end{equation}

\medskip \noindent where $R_{X}$ is the reward at time $t$ for each task, $\lambda$ is a scaling factor, $x_t$ and $x_{t-1}$ is the position of the agent at time $t$ and $t-1$ respectively, $x_0$ is the initial position of the agent, and $x_f$ is the position of the goal (i.e., end of maze for \navigation, goal object in \detection). The $w$ variables are non-zero when certain conditions are met. $w_g$ and $w_a$ indicates the reward/penalty given for reaching the goal and adversary, respectively. $w_c$ is the penalty for contacting a wall. In essence, in the \navigation task, the agent is incentivized to move from it's initial position as fast as possible. In the \detection and \tracking tasks, the agent is incentivized to navigate to the goal as quickly as it can. During training, $\lambda = 0.25$, $w_g = 1$, $w_a = -1$, and $w_c = -1$. Additionally, when an agent reaches the goal or adversary, the episode terminates.

\bigskip \noindent \textbf{Fitness function.} As compared to the reward function, the fitness function is used to evaluate the current performance of an agent. Where the reward function is used to inform the RL algorithm for it's weight optimization, the fitness function informs the evolutionary search algorithm for further selection and mutation. For each task, the fitness function $F_{X}$ in generation $g$ is identical to $R_{X}$, except with different weights to emphasize the relative performance difference between agents. During fitness evaluation, $\lambda = 1.5$, $w_g = 10$, $w_a = -10$, and $w_c = -2$. Instead of terminating when the goal or adversary is reached, in evaluation, the object is respawned and the agent continues to solve the task.

\bigskip \noindent \textbf{Evolution loop.} Evolving the full agent visual genotype, with a total of \small$>\sim$\normalsize$10^{20}$ possible combinations, necessitates ``intelligent'' optimization. The vast size of the search space alone means all combinations cannot be tested in a timely fashion. Strategically selecting morphologies that simultaneously explore new configurations and exploit previously gained knowledge is imperative to not waste resources on suboptimal solutions. We accomplish this intelligent search mechanism through the integration of evolutionary strategies (ES) \cite{salimans2017evolution}. ES is a broad optimization technique that is inspired by natural evolution and operates by iteratively refining a population of candidate solutions through processes such as mutation, selection, and adaptation. Unlike traditional genetic algorithms, ES emphasizes mutation over crossover and is particularly well-suited for optimizing continuous, high-dimensional spaces.

We use a population size of 16 agents, and evolve for 50 to 100 generations depending on the experiment. The specific ES algorithm we use is the Covariance Matrix Adaptation Evolution Strategy (CMA-ES) \cite{hansen2001completely}. CMA-ES is a variant of ES that adapts the mutation distribution based on the covariance matrix of the population. This adaptation allows for faster convergence and better exploration of the search space. We use the open-source implementation of CMA-ES provided by nevergrad \cite{nevergrad}. Hyperparameters for CMA-ES can be found in the Supplementary Material.

\bigskip \noindent \textbf{Agent phenotype.} An agent's phenotype is the physical manifestation of its genotype. The phenotype is the realized form that interacts within the environment that acquires and acts on observed stimuli. An agent in this work is represented as a fixed radius sphere with eyes facing outward and distributed uniformly along its equator. The agent is embodied therefore can control its  direction and speed using its underlying policy, which are used to actuate the joints to move the agent in the simulation environment. Our framework also allows for more complex dynamical systems and controller to be used as well. Lastly, in the case of the \tracking task, we assign computed action profiles to the goal and adversary to move to random locations within the environment. 

\label{sec:methods-genotype}
\bigskip \noindent \textbf{Agent's genotype.} The genotype encodes the instructions to create the agent's eye and cognitive system that learns task behavior. The vision genotype are further divided into three clusters that are mutated independently, and incorporate both continuous and discrete parameters. The clusters are: morphological, optical, and neural \Cref{fig:genotype}. Rather than modeling complex biological mechanisms like photoreceptor dynamics, we implement these genes to encode capture of light from a physics-based rendering perspective. We believe that this model captures the essential functional properties needed to evolve vision while remaining computationally tractable in modern embodied simulators. Our full encoding scheme is capable of representing approximately $10^{20}$ unique agent vision types.

\bigskip \noindent \textbf{Morphological genes.} The morphological genes defines properties used to spatially sample the environment, such as the number of eyes, their placement (determined by placement range), and field of view (FOV). We model the agent as a sphere of fixed radius of 0.2 units with eyes distributed uniformly along its equator. Thus, the placement of each eye is also dependent on both the number of eyes, and a placement range (i.e., the maximum angle from latitude 0$^\circ$) that eyes are uniformly distributed within. For instance, if an agent has 3 eyes and the placement range is 90$^\circ$, the eyes are placed at -45$^\circ$, 0$^\circ$, and 45$^\circ$, however, a placement range of 20$^\circ$ will result in three forward facing eyes at -10$^\circ$, 0$^\circ$, and 10$^\circ$. The orientation of each eye is determined by the normal vector from the agent's body. We assume bilateral symmetry, consistent with the observation that the overwhelming majority of animals have bilateral symmetry \cite{hollo2012manoeuvrability}. The FOV is a continuous integer value $[1^\circ, 100^\circ]$. We visualize the effect of sampling different morphological genes on agent's vision in \Cref{fig:plenoptic_num_eyes}, and \Cref{fig:plenoptic_resolution}.

\bigskip \noindent \textbf{Optical genes.} The optical genes describes how each eye interacts with incoming light in a physically plausible way. It encompasses a programmable Diffractive Optical Element, or a phase mask, that modulates the phase of the incoming light (represented as a $4 \times 4$ array with $\in [0, 1]$), refractive index ($ \eta \in [1.0, 2.0]$) and pupil radius, or aperture $a \in [0,1]$ that is scaled dynamically with sensor size as $r = a \times L$, where $L$ is the sensor size. Note that pupil size of $a=1$ results in an ``open eye'' as the pupil and sensor size are equal.  We use continuous parameters for phase mask, pupil radius, and refractive index and then upsample the phase mask to a size of $51,51$ to generate smoother PSFs. The optical gene can also be disabled in which case a rasterization-based imaging model is used to create visual stimuli for the agent; this model is analogous to an eye with a perfect lens, i.e no blur. Our framework doesn't limit the size of the phase mask, however, we choose $4 \times 4$ to make the problem computationally tractable. We visualize the effect of sampling different optical genes on agent's vision in \Cref{fig:plenoptic_aperture}, \Cref{fig:plenoptic_eta} and \Cref{fig:plenoptic_phase}.

\bigskip \noindent \textbf{Neural genes.} The neural gene determines the learning capacity of the agents by encoding properties of the agent's neural network, such as temporal memory size and number of neurons. The memory size represents the number of historical frames relative to the current frame the agent has access to; for instance, a memory size of five means the agent's visual stimuli is a flattened vector each composed of the current frame and the previous five frames. For \Cref{sec:results-neural}T we sample different neural genes from an underlying neural network architecture that has two identically sized hidden layers. The number of neurons in the hidden layers is an integer mutation parameter that can be between 1 and 512. Although we don't present experiments in this work evolving the neural network architecture, it is possible to mutate the underlying architecture (i.e., add layers, change activation functions, etc.) in our framework.

\bigskip \noindent \textbf{Mutation operators.} The agent's genotype is designed with specific mutation operators for each parameter type. For continuous parameters (such as FOV, phase mask values, pupil radius scaling factor, and refractive index), mutations occur through gaussian perturbation within their defined ranges. For discrete integer parameters (such as number of eyes and number of neurons), mutations increment or decrement the current value while respecting the parameter bounds. Bilateral symmetry is preserved during all morphological mutations.

\bigskip \noindent \textbf{Experimental control.} In each experiment, we enable specific mutations to isolate and study specific aspects of vision evolution. This controlled approach allows us to systematically investigate the evolution of different visual traits.

\begin{figure}[t]
    \centering
    \includegraphics[width=0.9\textwidth]{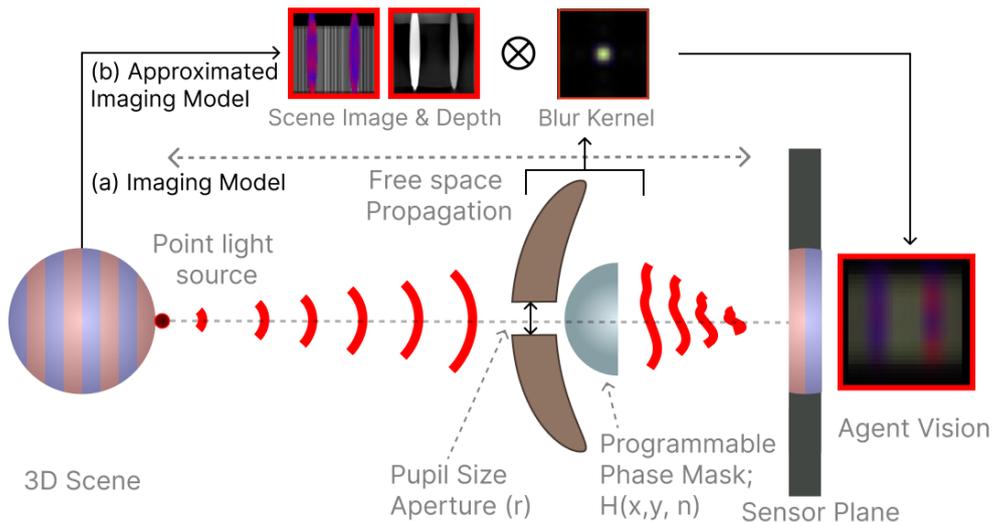}
    \caption{\textbf{Imaging Model.} Our simulation implements both wave and geometric optics using a OpenGL \cite{woo1999opengl} \textbf{(a)} Our scene imaging model shows how a depth dependent blur kernel is derived: light from a point source in the 3D scene propagates through free space, passing through a pupil plane which is composed of (1) an aperture with a variable aperture radius ($r$) and (2) a programmable phase mask with height map $H(x,y)$ and refractive index $n$, before forming an image on the sensor plane. \textbf{(b)} The approximated model uses a 2D convolution between the scene, a depth map, and a single blur kernel using the far-field approximation (i.e., the Point Spread Function) for computational efficiency.}
    \label{fig:methods-imaging}
\end{figure}

\bigskip \noindent \textbf{Imaging model.} To produce environmental pressures based on physics of light transport, we use a physically-based imaging model for the agent's eye. This ensures that agent must evolve their optical genes to contend with physical laws that limit visual systems. Thus, our imaging model consists of a programmable lens described by a phase mask, refractive index, and amplitude modulating (aperture) elements described by a pupil/aperture radius. The  agent's retina is emulated by a discretized pixel at the sensor plane at a focal length distance away from the pupil (\Cref{fig:methods-imaging}).

All imaging systems capture the scene as an optically encoded image on to the sensor plane. These optical encodings are commonly referred to as the blur or point spread function (PSF), and are dependent on the phase and amplitude of the pupil function along with wavelength and depth of the scene point. We follow the wave propagation model described in \cite{tasneem2022learning,chang2019deep, chang2019deepopticsmonoculardepth} to estimate the  PSF. Since this model will be applied per eye per step of the agent there is an inherent tradeoff between speed of rendering and accuracy of the imaging model. Thus, we assume depth in-dependent PSFs to limit the number of depth-convolutions, to one convolution, per eye. Additionally, to reduce the number of evolvable parameters for the amplitude mask we also assume a circular amplitude mask parametrized by a single value (pupil radius). Our framework is not limited to these assumptions. 

Given a point light source at a distance $z$ and the pupil function $P(x,y) = A(x,y) \exp(i \times t_{\phi}(x, y))$ (\Cref{fig:methods-imaging}) the response of the agent's eye can be measure by the PSF. The PSF at the sensor plane $s$ distance away from the pupil plane is described as:

\begin{equation}
    PSF_{\lambda,_z}(x', y') = \left| \mathcal{F}^{-1} \left\{ \mathcal{F} \left\{ P(x, y) U_{in}(x, y) \right\} H_s(f_x, f_y) \right\} \right|^2,
\label{eq:psf}
\end{equation}

\medskip \noindent where $H_s(\cdot)$ represents the field propagation transfer function \cite{goodman2005introduction} for distance $s$ with $(f_x, f_y)$ as the spatial frequencies given as

\begin{equation}
    H_s(f_x, f_y) = \exp\left[ iks \sqrt{1 - (\lambda f_x)^2 - (\lambda f_y)^2} \right];
\end{equation}

\medskip \noindent where \( k = \frac{2\pi}{\lambda} \) is the wavenumber; $U_{in}(x,y)$ denotes the complex-valued wave field immediately before the lens which for a point light source is given as
\begin{equation}
    U_{in}(x, y) = \exp\left(ik\sqrt{x^2 + y^2 + z^2}\right);
\end{equation}

\medskip \noindent $\mathcal{F} \{\cdot\}$ is the 2D Fourier transform; $(x', y')$ are the spatial coordinates on the camera plane, and $(x, y)$ are the coordinates on the lens plane.

\noindent The pupil function pupil function $P(x,y) = A \exp(i \times t_{\phi}(x, y))$, contains an amplitude modulate function, $A(x,y)$ and a phase modulation function $t_{\phi}(x, y)$. To keep the number of evolvable parameters small, we assume a circular amplitude mask: $(x - \frac{W}{2})^2 + (y - \frac{H}{2})^2 = r^2$ where $r \in [0,1]$ and is an mutated parameter. The phase modulation function is represented by $t_{\phi}(x, y) = e^{i \frac{2\pi}{\lambda} \phi(x, y)}$ in \Cref{eq:psf} is generated by the lens surface profile $\phi(x,y)$ which in our case is a mutating square 2D phase-mask array of size 16, where $\phi(x,y) \in [0, 1]\}$. 

\begin{equation}
    U_{in}(x, y) = \exp\left(ik\sqrt{x^2 + y^2 + z^2}\right);
\end{equation}

\medskip \noindent Finally, our agent's image formation follows a shift-invariant convolution of the image and the depth-independent PSF to yield the final image, $I_{\ell}$, perceived by the agent. 
\begin{equation}
    I_{\ell} = (\mathcal{S}_{\ell}(H_{\ell} \ast X_{\ell}) \times r^2) + N_{\ell},
\end{equation}

\medskip \noindent where the sub-index $\ell$ denotes the color channel; $X_{\ell} \in \mathbb{R}^{w \times h}_{+}$ represents the underlying scene with $w \times h$ pixels; $H_{\ell}$ represents the discretized version of the PSF in \Cref{eq:psf}; $N_{\ell} \in \mathbb{R}^{w \times h}$ denotes the Gaussian noise in the sensor; $\mathcal{S}_{\ell}(\cdot): \mathbb{R}^{w \times h} \rightarrow \mathbb{R}^{w \times h}$ is the camera response function, modeled as a linear operator, multiplication by $r^2$ denotes changing light throughput onto the pixel area as a result of the aperture and $\ast$ denotes the 2D convolution operation. The light throughput falls quadratically because light falls onto a sensor plane which has finite area in x and y.

In practice, the discretized version of the PSF (H, W in $A(x,y)$ and $t_{\phi}(x, y)$) is of size $(H+1, W+1)$ where $(H, W)$ is the resolution of the agent's eye. This is an explicit choice to make the PSF larger than the image to enable a full blur on the eye when the aperture is fully open. To make sure that angular scene resolution is maintained, the scene image $X_{\ell}$ is rendered by padding $I_{\ell}$ of size $(H, W)$ to $\left( H + (2*\frac{H+1}{2}), W + (2*\frac{W+1}{2}) \right)$. 

This ensures that similar to a real eye, the corner photoreceptors also accumulate light from outer regions directly due to the aperture size. This also means that closing the aperture also helps with reducing the total effective field of view of the agent's eye and less blur, which is how agents control the blur in Phase I experiment in \Cref{sec:results-optics}. For pinhole eyes, the field-of-view becomes equivalent to the encoded fov in the agent's morphological gene as the blur kernel is very small.


\bigskip \noindent \textbf{Simulated environment.} This framework is built on top of a the MuJoCo physics engine \cite{mujoco} and within a gymnasium-style \cite{towers2024gymnasium} setup. The underlying dynamics of each agent is governed by the MuJoCo physics and images are rendered via a rasterization pipeline using the builtin OpenGL renderer. We customize this environment with the physically-based rendering model. 

In this work, we differentiate between a visual task and an environment. A visual task is the specific goal an agent is trying to achieve defined by the reward function. The environment is the physical space in which the agent is placed, and so can contain multiple tasks. For instance, the same environment is used to train agents on both the detection and tracking tasks; the only difference is the reward function and the movement of the goal/adversary. In addition to the physical positioning of objects in an environment, the textures, light, colors, etc. can be modified to create a diverse set of various environments.

Each environment also have walls, which can are organized as boundary or in a maze-like configuration. These walls are rigid and contactable, and provide barriers where the agent cannot move escape. We can also add any textures on these walls. While we only train a single agent that is evolved and trained in an environment, we also have non-trainable agents. Non-trainable agents have a fixed action policy and have privileged information about the environment. For example, a non-trainable goal object in the \tracking task has fixed policy that randomly samples actions to move in the environment. 

\clearpage
\begin{appendices}
\section{Analysis of Evolved Agents}
\label{app:lensing_analysis}

\begin{figure}[ht!]
    \centering
    \includegraphics[width=\textwidth]{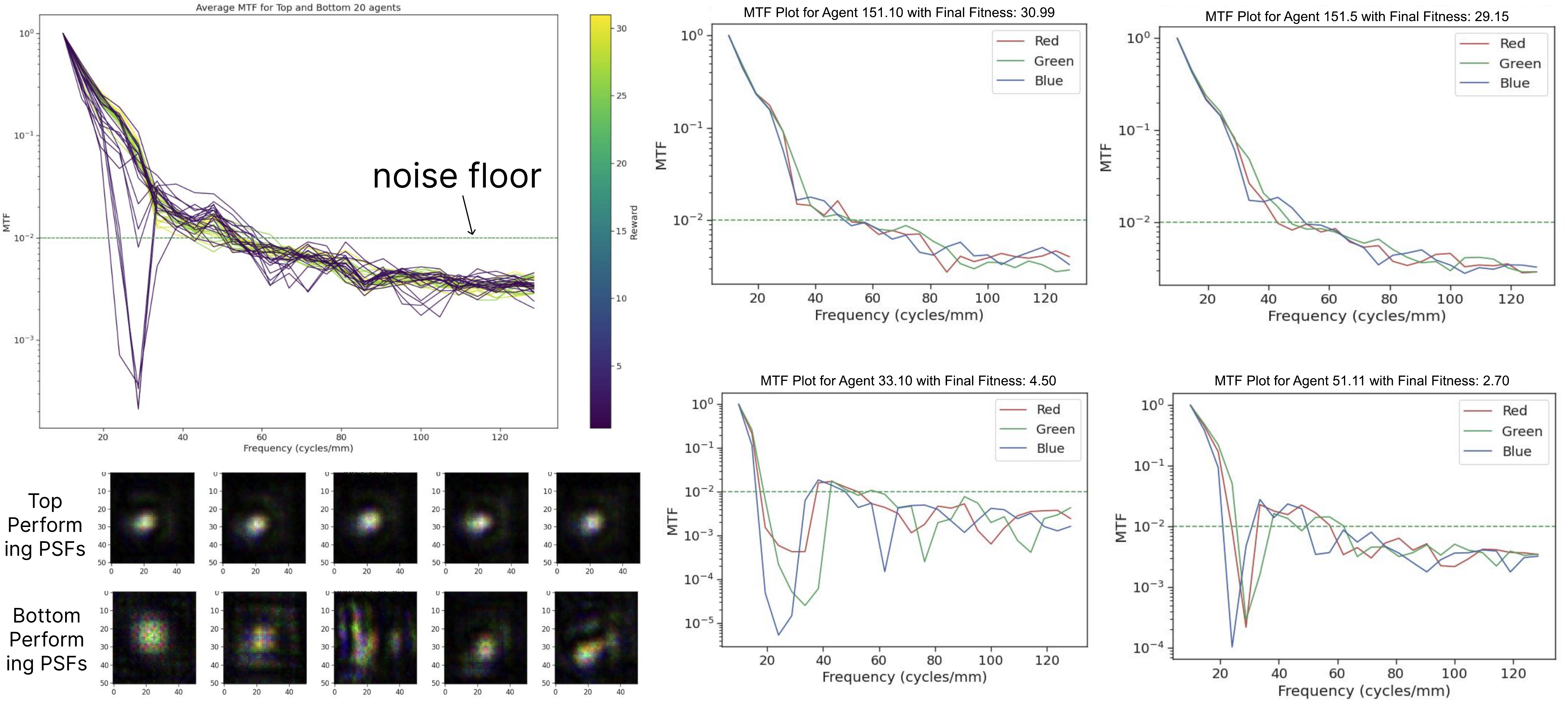}
    \caption{\textbf{Modulation Transfer Function (MTF) analysis of evolved vision systems:} We analyze the spatial frequency response of evolved vision systems using MTF curves, which quantify how well different spatial frequencies are preserved. Left: Average MTF curves for top 25 and bottom 25 performing agents, colored by reward. The horizontal dashed line indicates the noise floor, below which spatial information becomes unreliable. Bottom: Point Spread Functions (PSFs) for the best and worst performing agents, showing the characteristic light distribution patterns. Top Performing Agents develop compact and symmetric PSFs even though we don't enforce any symmetry in our setup. Right: Individual MTF curves for agents at different evolutionary stages (Generation 33, 51, and 151) and performance levels. Early generations (Gen 32, 51) show erratic frequency responses with significant dips, indicating poor optical performance. By Generation 151, high-performing agents develop smooth MTF curves that maintain good contrast above the noise floor up to ~40 cycles/mm, demonstrating evolution of effective lens-based vision systems. The RGB channels show similar responses, suggesting achromatic optimization of the optical system. Note that the dispersion in the PSFs will be clearer in the pdf version compared to the print.}
    \label{fig:supp_mtf}
\end{figure}

\begin{figure}[ht!]
    \centering
    \includegraphics[width=\textwidth]{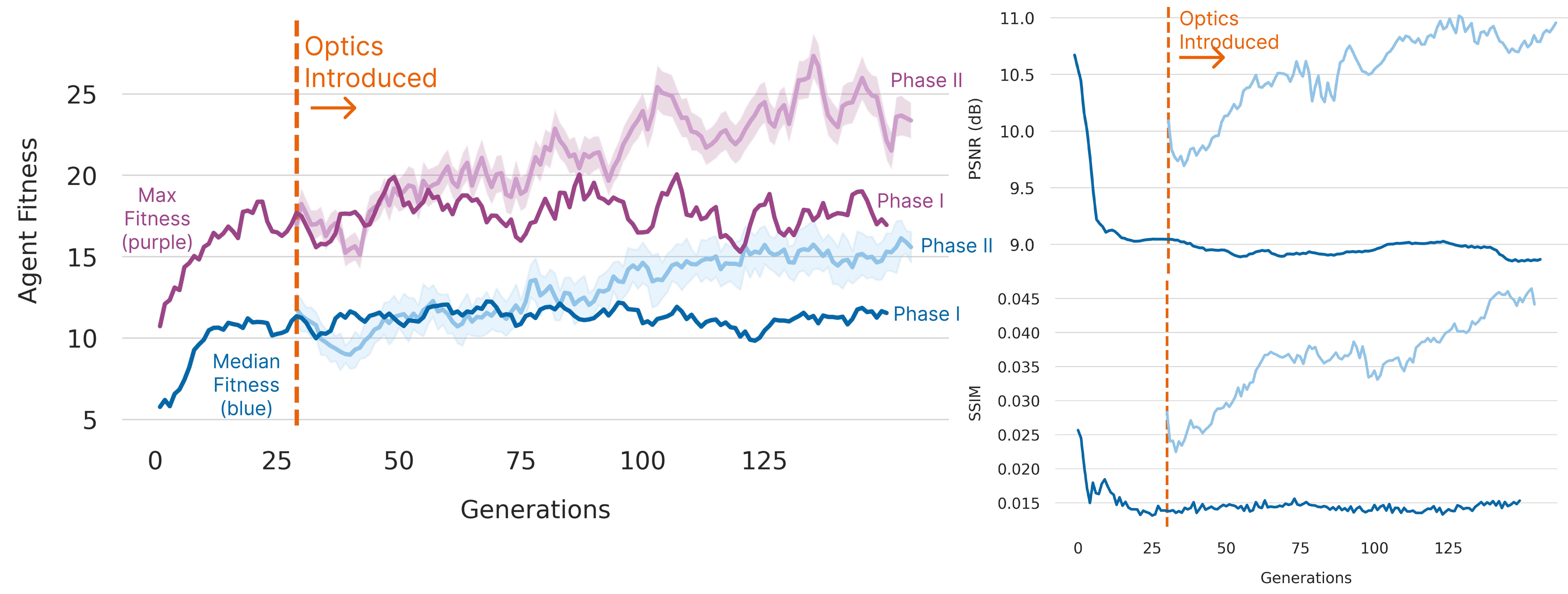}
    \caption{\textbf{Agent Fitness, PSNR (dB) and SSIM over generations:} We show additional graphs from Section 2.4 that show steady increase in agent fitness. The main plot tracks both maximum (purple) and median (blue) agent fitness for Phase I (solid) and Phase II (opaque). In Phase I, median fitness saturates at 10 indicating no food detection, while in Phase II, median fitness exceeds 18 by generation 130, demonstrating evolved capability for reliable food detection. The maximum fitness of >25 in Phase II indicates multiple successful detections, whereas Phase I's maximum fitness shows only occasional random successes. To compute SSIM and PSNR we render a rastered version of the image using the pinhole camera model as the reference image. For PSNR we show a substantial increase from the pinhole eyes. This shows that lensing significantly improves the signal-to-noise tradeoff that we discussed in Section 2.4. We can also compare the SSIM between Phase I and II. Unlike PSNR, SSIM is a perception-based model that considers changes in structural information between reference and target images. The SSIM increases are little as it relies on pixel wise calculations but the trend demonstrates that lens-based eyes better preserve image structure while maintaining higher light collection compared to pinhole eyes, enabling more reliable discrimination between similar visual features.}
    \label{fig:supp_fitness}
\end{figure}

\begin{figure}[ht!]
    \centering
    \includegraphics[width=0.75\textwidth]{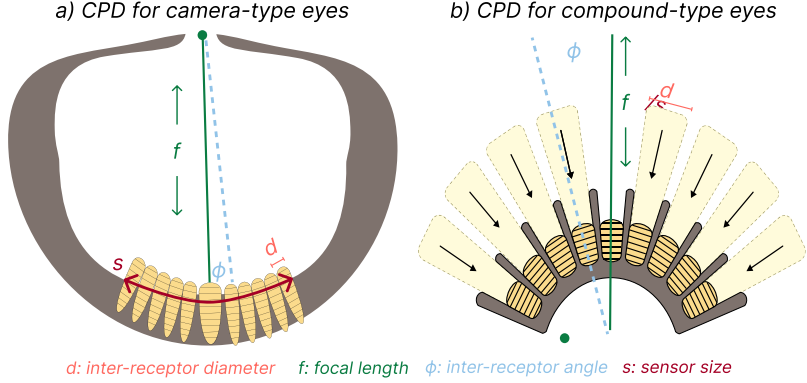}
    \caption{\textbf{Comparative analysis of cycles per degree (CPD) calculations for agents with different eye morphologies.} We show the schematic representation of two distinct eye architectures and their corresponding CPD calculation methods. (a) An example camera-type eye configuration is shown with a single eye with 11 photoreceptors, characterized by focal length ($f$), inter-receptor diameter ($d$), inter-receptor angle ($\phi$), and sensor size ($s$). (b) An example Compound-type eye configuration with 9 individual eyes, each containing a single photoreceptor. The field of view (FOV) is determined by the relationship between sensor size and focal length, while the range of placement of eyes denoted as longitudinal range, lon\_range, is calculated as the product of sensor size and number of eyes. For our agents, we compute CPD using the minimum ratio of longitudinal range to the number of eyes minus one, and the field of view to resolution ratio ($\text{FOV}/\text{resolution}$). Note that if that the compound eye had more than one photreceptor $d$ would be a per-photreceptor measurement and $s$ would remain the same, however, in this configutation they are the same since we are showing a simpler case of one photreceptor per eye.}
    \label{fig:supp_acuity}
\end{figure}

Our analysis quantifies the optical performance of evolved vision systems using three metrics. The Point Spread Function (PSF) represents the system's response to a point source of light - how a perfect point gets ``spread out" by the optical system. In \Cref{fig:supp_mtf}, high-performing agents develop compact, symmetric PSFs indicating precise light focusing, while poor performers show diffuse, irregular patterns suggesting inefficient light management. A perfect PSF would appear as an infinitesimally small point, while real optical systems produce some degree of spread due to diffraction and optical imperfections. \newline

\noindent \textbf{MTF for Spatial Precision Analysis} The Modulation Transfer Function (MTF), mathematically derived as the Fourier transform of the PSF, quantifies how well different spatial frequencies are preserved by the optical system. On the MTF plots in \Cref{fig:supp_mtf}, the y-axis represents contrast preservation (from 0 to 1) while the x-axis shows spatial frequency in cycles/mm. The area above the noise floor $(~10^{-2})$ represents useful spatial information - frequencies where the signal can be reliably distinguished from noise. Early-generation agents show erratic MTF curves with sharp dips below this noise floor, while later-generation agents maintain smooth curves above the noise floor up to ~40 cycles/mm. \newline

\noindent \textbf{Image Quality Metric.} These observations led us to develop an Image Quality metric that multiplies two factors: (1) the area under the MTF curve above the noise floor (marked in \Cref{fig:supp_mtf} by the dashed noise floor line), representing spatial precision, and (2) light throughput, which decreases quadratically with pupil radius. This metric captures the trade-off between spatial precision and light collection. While pinhole eyes can achieve good MTF performance, their limited light throughput constrains their overall image quality. Lens-based eyes resolve this trade-off by maintaining strong MTF performance while allowing larger apertures for better light collection. \newline

\noindent \textbf{Quantifying Signal-to-Noise Ratio in Evolved Agents.} The PSNR and SSIM curves in \Cref{fig:supp_fitness} reveal limitations of these conventional metrics. Initially, with fully open apertures, both metrics show high values because extreme blur acts as a noise-reducing low-pass filter. However, this blurred vision makes the detection task impossible, resulting in low agent fitness. As apertures begin to close (forming pinhole eyes), PSNR and SSIM decrease as the system preserves more high-frequency information but with increased noise due to limited light collection. When lensing is enabled at generation 30, we observe steady improvement in both metrics as the system evolves the ability to maintain high-frequency detail while collecting sufficient light. \newline

\noindent \textbf{Analyzing fitness trajectories for DETECTION task.} In the DETECTION Task a fitness of $>15$ means that the agent has correctly detected food at least once (each food detection has a $+10$ fitness score), and a fitness of 25 means that this has happened at least twice. In \Cref{fig:supp_fitness}, we plot the median fitness for phase I which saturates at a fitness score of 10 which results in the agent not detecting any food. However, median fitness for Phase II by generation 130 is $>18$ which shows that evolution has discovered a reliable way to continuously detect food from 2 poison objects. Moreover, the best agents can do it multiple times as the max fitness score for Phase II is $>25$. Comparatively, agents with max fitness for Phase I (solid purple line) detect food at least once which is mostly a result of randomness and getting lucky. \newline

\noindent \textbf{Randomness in Agent Behavior.} Notably, some agents with excellent optical properties (high MTF and light throughput) show slightly lower rewards due to the stochastic nature of reinforcement learning. This variance in reward despite similar optical quality suggests that the relationship between optical performance and task success is not purely deterministic - better vision enables but does not guarantee better task performance.\newline

\begin{equation}\label{eq:supp-cpd}
\text{CPD} = \frac{1}{2 \cdot \min\!\Big (\frac{\mathrm{lon\_range}}{\mathrm{num\_eyes} - 1}, \frac{\mathrm{fov}}{\mathrm{resolution}} \Big)}
\end{equation}

\begin{equation}\label{eq:supp-fov}
\mathrm{FOV} = 2 \arctan\!\Big(\frac{\mathrm{sensor\_size}}{2 \times \mathrm{focal\_length}}\Big)
\end{equation}

\begin{equation}\label{eq:supp-lon-range}
\mathrm{lon\_range} = \mathrm{sensor\_size} \times \mathrm{num\_eyes}
\end{equation}

\noindent \textbf{Quantifying Agent Vision through CPD.} We also quantify an agent's morphological and optical genotype in cycles per degree (CPD). The cycles per degree is a measure of the spatial frequency observable to the imaging system; a higher CPD value corresponds to a better ability to resolve fine spatial details and distinguish closely spaced features in the visual scene \cite{caves2018visual}. CPD is also a commonly used metric to measure visual capabilities in real-life animals \cite{caves2018visual}. \Cref{eq:supp-cpd}, \Cref{eq:supp-fov}, and \Cref{eq:supp-lon-range} show are used to calculate CPD for our agents.

\begin{figure}[ht!]
    \centering
    \includegraphics[width=\textwidth]{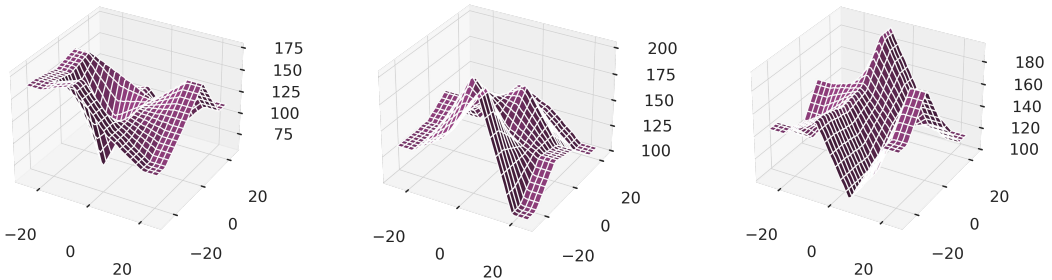}
    \caption{\textbf{Comparison of evolved optical elements between best and worst performing agents.} Three-dimensional surface plots showing the optical response patterns for the top six (upper panel) and bottom six (lower panel) performing agents. Best performing agents (F=24.4-31.0) exhibit well-defined, singular peak formations with smooth gradients, while worst performing agents (F=0.7-3.4) display irregular, multi-peaked patterns with abrupt transitions. Each plot represents a unique evolved optical configuration denoted by its agent identifier (A) and corresponding fitness score (F). }
    \label{fig:supp_optical_elements}
\end{figure}

\begin{figure}[ht!]
    \centering
    \includegraphics[width=\textwidth]{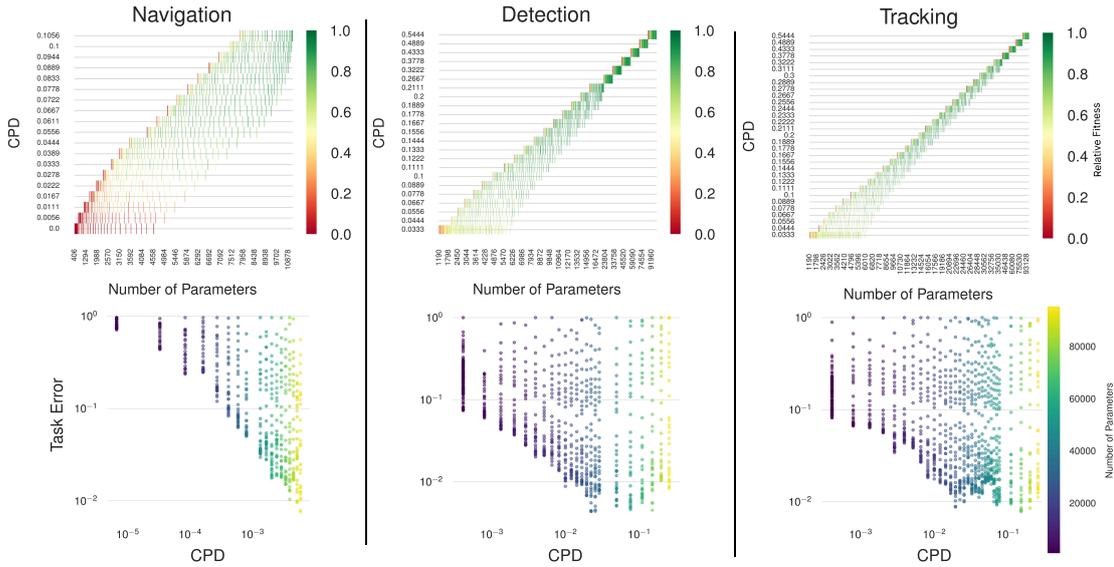}
    \caption{\textbf{Dense parameter analysis revealing task-specific relationships between sensory acuity and neural processing.} \textbf{Top:} Performance visualization showing individual trials (vertical lines) across CPD values and network sizes for navigation (left), detection (middle), and tracking (right). \textbf{Bottom:} Corresponding scatter plots with log-scaled axes demonstrate how error rates vary with CPD for different network sizes (indicated by color intensity). The distinct patterns across tasks support our findings about task-dependent scaling relationships between sensory and neural resources.}
    \label{fig:supp_neural}
\end{figure}

\section{Genotype and its relation to the Plenoptic Design Space}
\label{app:design-space}

The vision genotype of the agent can be understood as operating on the plenoptic function, which describes the complete flow of light in a scene \cite{adelson_plenoptic}. We can conceptually think about biological vision evolution as directly evolving to capture differnet dimensions of the plenoptic function. Our current implementation demonstrates that computational evolution can also sample from the Plenoptic function \cite{adelson_plenoptic}: we allowing evolution to explore a subset of plenoptic dimensions such as placement, orientaotin, optical constraints, movement of the agent etc. We believe this  framework can naturally extend to encompass the full plenoptic representation of light - including spectral sensitivity, polarization detection, and varied spatiotemporal resolutions. Just as our computational experiments have shown evolution discovering diverse and creative solutions within a limited set of visual parameters, expanding the genotype to sample from the complete plenoptic dimensions would enable the discovery of even more sophisticated visual systems, analogous to those found in nature. For instance, the mantis shrimp (stomatopods) evolved 16 different photoreceptor types that can detect both linear and circular polarized light \cite{marshall1988unique}, while jumping spiders (Salticidae: Dendryphantinae) developed a unique combination that provide both high acuity and wide-field motion detection \cite{land1958structure}. Our genotype enables co-evolution of eyes, neural circuitry and subsequent behavior (learned through reinforcement learning) and provides a unified way to think about vision evolution as a creative optimization process operating directly on the fundamental properties of light. As we expand the genptype of the available plenoptic dimensions, we expect to see the emergence of increasingly sophisticated and novel visual systems that may parallel, or even exceed, the remarkable diversity found in biological evolution.

\section{Acuity-Neural Processing Trade-offs and Task-Specific Scaling}

In our framework, we systematically explore how visual task performance emerges from the interplay of three key components. The first component is the eye's physical characteristics, measured in cycles per degree (CPD), which determines the ability to resolve spatial detail. The second is neural capacity, where we vary the number of parameters in the vision-processing layers.

Our parameter sweep reveals emergent power law scaling relationships between sensory acuity and neural capacity \Cref{fig:supp_neural}. The relative fitness plots (top row) demonstrate that navigation achieves high performance ($>$0.8) at lower CPDs ($~$0.05) with modest neural capacity ($~$8000 parameters). Detection and tracking tasks show a distinct scaling pattern, requiring both higher CPDs ($>$0.3) and larger networks ($>$40,000 parameters) for comparable fitness levels.

The error plots (bottom row) reveal fundamental constraints in how these capabilities emerge. At fixed CPD values, increasing neural capacity follows characteristic power law improvements until hitting task-specific performance ceilings. These ceilings are particularly evident in the scattered error distributions, where higher CPDs enable lower minimum error rates across all tasks. This demonstrates that poor visual acuity creates a fundamental bottleneck that cannot be overcome by simply scaling neural capacity. Notably, detection and tracking display continuous improvements in error rates as both CPD and network size increase, suggesting these tasks benefit from simultaneous scaling of both sensory and neural resources.

These computational scaling relationships emerged spontaneously through evolution in our framework, revealing how physical constraints in sensory acuity interact with neural processing capacity to shape task performance. The distinct scaling patterns across tasks, particularly the earlier saturation in navigation compared to detection and tracking, suggest a natural hierarchy in the visual processing demands of different behaviors \cite{nilsson_Visual_Roles}. This emergent relationship between sensory hardware and neural processing mirrors both biological evolution \cite{2018AllometryTS, venditti2024co} and contemporary artificial intelligence scaling laws \cite{kaplan2020scaling, hoffmann2022trainingcomputeoptimallargelanguage, hestness2017deeplearningscalingpredictable}.

\begin{figure}[ht!]
    \centering
    \includegraphics[width=\textwidth]{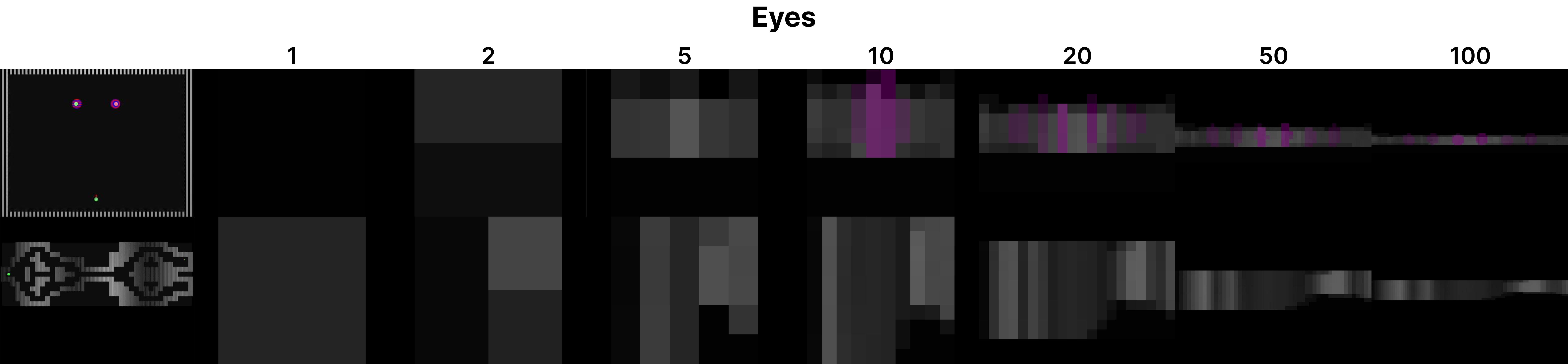}
    \caption{\textbf{Sampling greater number of eyes by modifications to the Morphological Gene:} We display our agent's vision captured with progressively more number of eyes. Top row shows progressively increasing eyes allows for larger FOV of the scene and creates multiple copies of the spheres from slightly different perspectives in each eye. The bottom row shows that for the navigation task number of eyes allows the agent to see different parts of the wall which it uses to orient itself against wall collisions.}
    \label{fig:plenoptic_num_eyes}
\end{figure}

\begin{figure}[ht!]
    \centering
    \includegraphics[width=\textwidth]{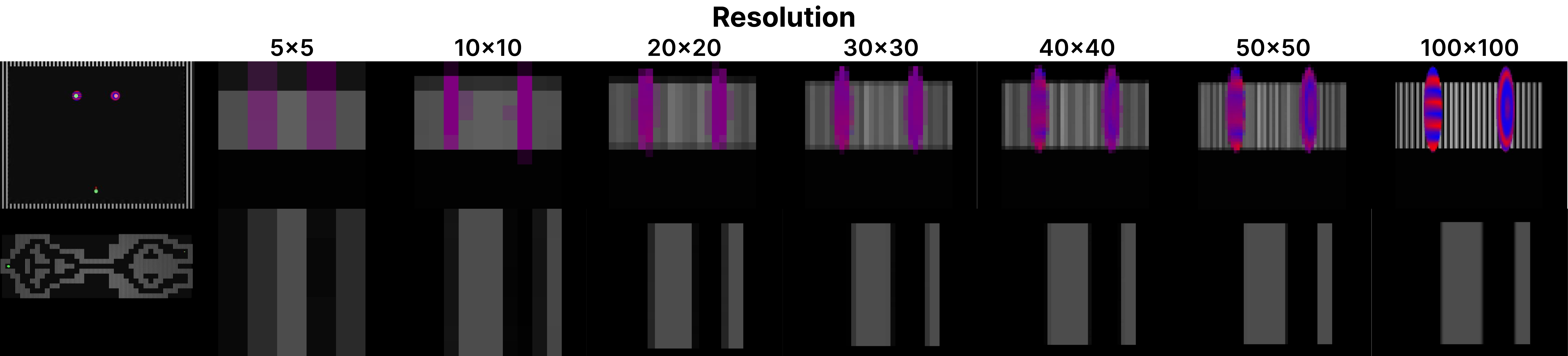}
    \caption{\textbf{Sampling larger resolutions by modifications to the Optical Gene:} We display our agent's vision captured with progressively larger resolutions. Top row shows progressively increasing resolution resolved the difference between food and poison which can be differentiated with the orientation of the stripes. The bottom row shows that for the navigation task resolution helps resolve the stripes on the wall.}
    \label{fig:plenoptic_resolution}
\end{figure}

\begin{figure}[ht!]
    \centering
    \includegraphics[width=\textwidth]{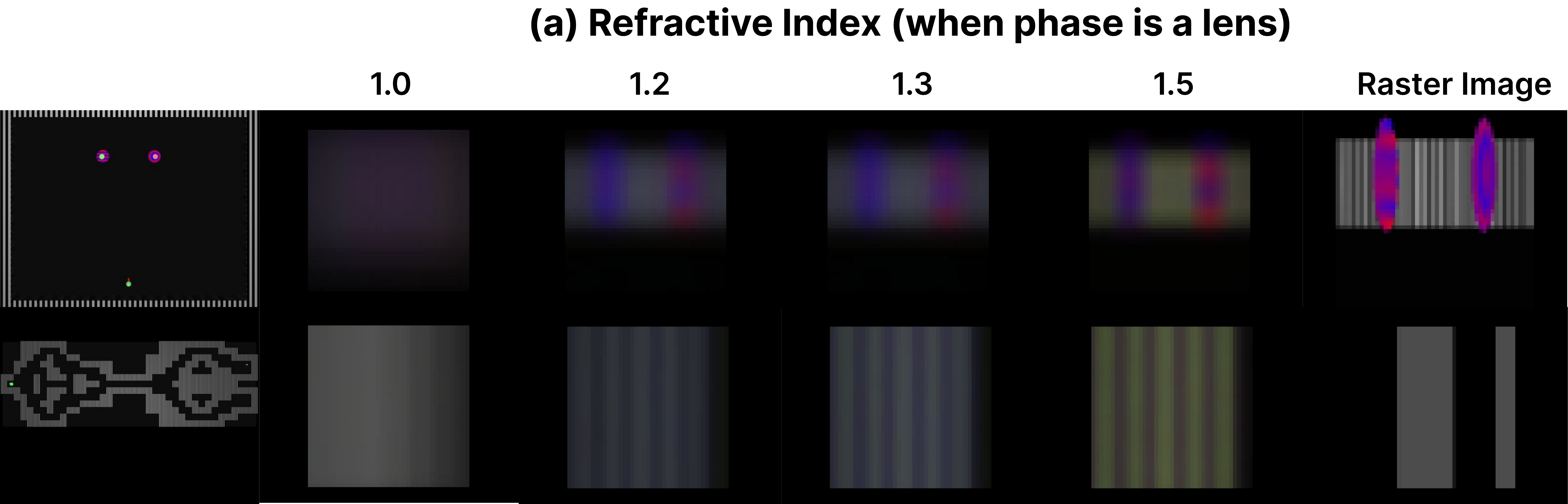}
    \caption{\textbf{Sampling refractive indices in the Optical Gene:} We illustrate examples of sampled refractive indices within the Optical Gene for Detection (top row) and Navigation(bottom row) tasks. For a fixed phase mask (a perfect lens) increases in refractive index causes increase sharper images.}
    \label{fig:plenoptic_eta}
\end{figure}

\begin{figure}[ht!]
    \centering
    \includegraphics[width=\textwidth]{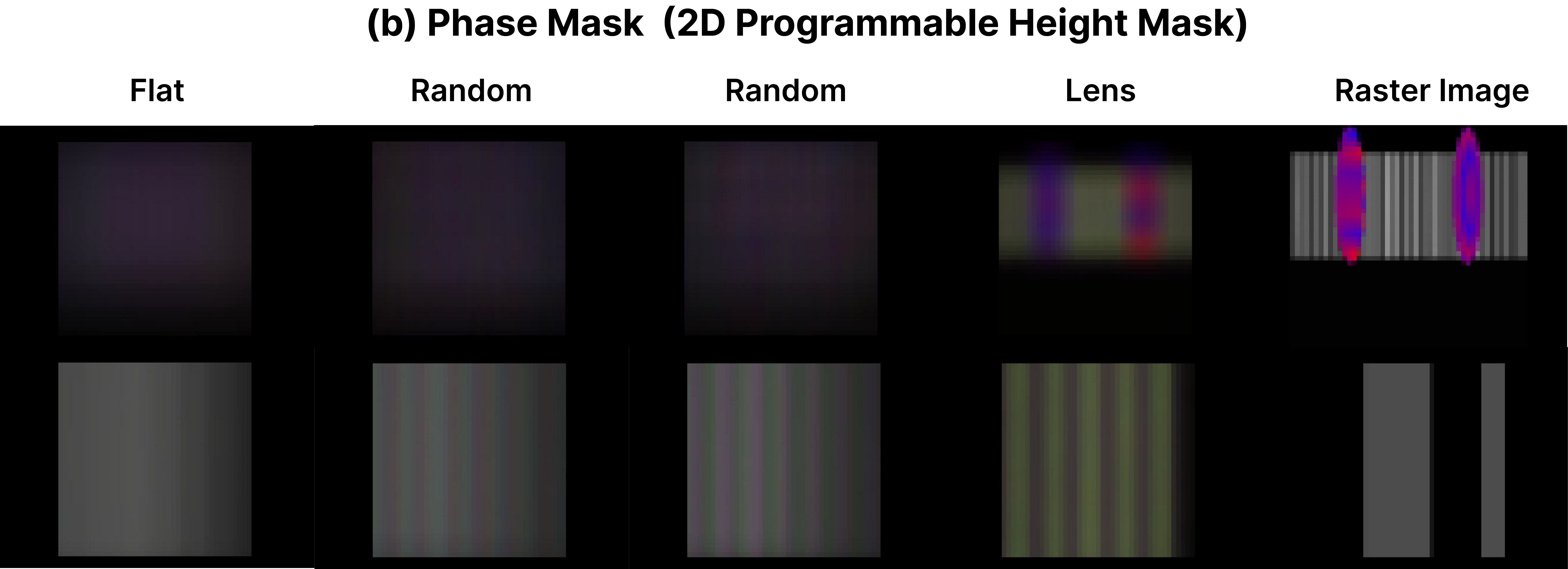}
    \caption{\textbf{Sampling optical elements in the Optical Gene:} We illustrate examples of sampled optical elements (phase masks) within the Optical Gene for Detection (top row) and Navigation(bottom row) tasks. The figure shows Flat, and 2 Randomly samples phase masks which shows the complexity of the design space. These visualizations also demonstrate that while using a lens is a major innovation in eye design, creating a focused lens is a hard problem that evolution solved really well.}
    \label{fig:plenoptic_phase}
\end{figure}

\begin{figure}[ht!]
    \centering
    \includegraphics[width=\textwidth]{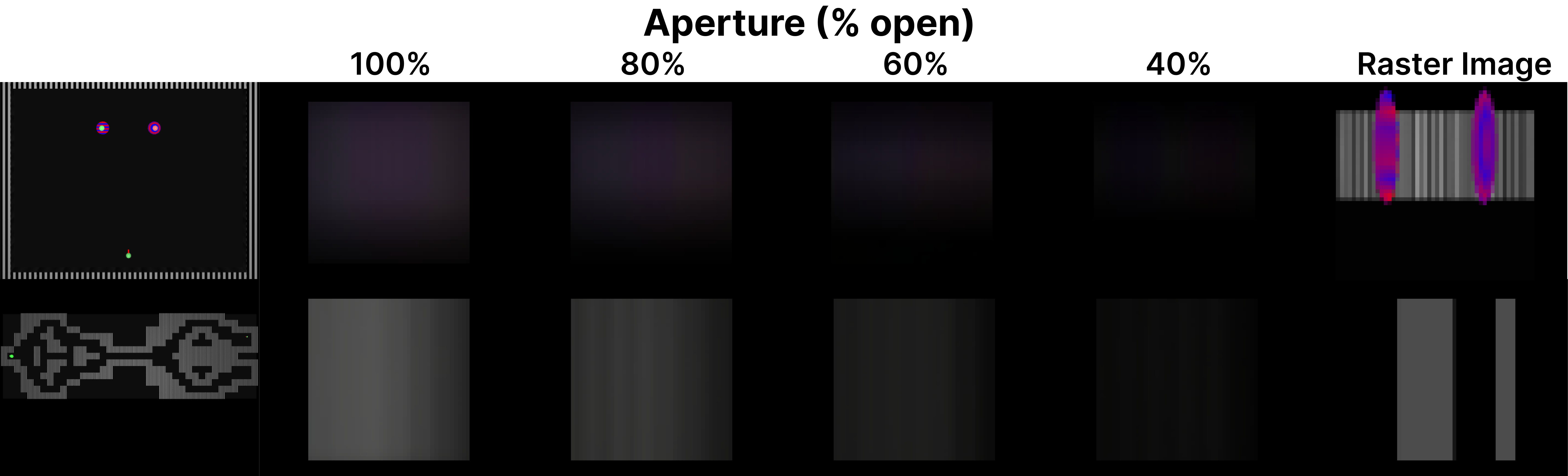}
    \caption{\textbf{Sampling Different Aperture by modifications to the Optical Gene:} We display our agent's vision captured with progressively smaller apertures, demonstrating how reducing the aperture size leads to increased image sharpness. However, as the aperture closes, the signal strength decreases quadratically with its radius, leading to higher noise levels. The balance between sharpness and noise is a critical factor for agents to successfully complete their visuomotor tasks.}
    \label{fig:plenoptic_aperture}
\end{figure}

For temporal performance, we find that performance saturates beyond 10 frames across all configurations. This is particularly evident in tracking tasks, where agents are incentivized to complete objectives quickly, typically achieving success in under 10 frames. This reveals an optimal balance between temporal information and computational efficiency that varies by task complexity.






\end{appendices}


\clearpage
\bibliography{sn-bibliography}


\end{document}